\documentclass[preprint,1p,times]{elsarticle}

\usepackage{amsmath}
\usepackage{amssymb}
\usepackage{booktabs}
\usepackage{longtable}
\usepackage{array}
\usepackage{calc}
\usepackage{graphicx}
\usepackage{float}
\usepackage{lscape}
\usepackage{tikz}
\usepackage{xcolor}
\usepackage[colorlinks,citecolor=blue,linkcolor=blue,urlcolor=blue]{hyperref}
\usetikzlibrary{arrows.meta, positioning, calc}

\makeatletter
\providecommand{\doi}[1]{\begingroup\catcode`\_=12 \catcode`\#=12 \catcode`\&=12 \@doi{#1}\endgroup}
\providecommand{\@doi}[1]{\href{https://doi.org/#1}{\nolinkurl{doi:#1}}}
\makeatother

\definecolor{lv0}{RGB}{34,139,34}
\definecolor{lv1}{RGB}{124,179,66}
\definecolor{lv2}{RGB}{255,235,59}
\definecolor{lv3}{RGB}{255,167,38}
\definecolor{lv4}{RGB}{244,67,54}
\definecolor{lv4b}{RGB}{183,28,28}
\definecolor{lv5}{RGB}{136,14,79}
\definecolor{lv6}{RGB}{50,50,50}

\journal{Information and Computation}

\begin{document}

\begin{frontmatter}

\title{The Generation-Recognition Asymmetry: Six Dimensions of a Fundamental Divide in Formal Language Theory}

\author[ind]{Romain Peyrichou\corref{cor1}}
\cortext[cor1]{Corresponding author}
\ead{research@roomi-fields.com}
\address[ind]{Independent Researcher}

\begin{abstract}
Every formal grammar defines a language and can in principle be used in three ways: to \emph{generate} strings (production), to \emph{recognize} them (parsing, analysis), or --- given only examples --- to \emph{infer} the grammar itself (grammar induction). Generation and recognition are extensionally equivalent --- they characterize the same set --- but operationally asymmetric along several independent axes. Inference is a qualitatively harder problem: it has no access to a known grammar --- the grammar is precisely what it seeks to discover. Despite the centrality of this triad to compiler design, natural language processing, and formal language theory, and although the comprehension/production/acquisition triad has been the subject of unified frameworks in psycholinguistics (notably the \emph{P-chain} of \citealt{dell2014pchain}), no study in formal language theory has, to our knowledge, treated this asymmetry as a structural multidimensional framework.

This paper focuses on the generation-recognition duality, incorporating inference as one of six dimensions of the asymmetry (D5: inference as the extreme case of recognition under decreasing grammatical knowledge). We identify \textbf{six dimensions} along which generation and recognition diverge: computational complexity, ambiguity, directionality, information availability, grammar inference, and temporality. We show that the common characterization ``generation is easy, parsing is hard'' is misleading: unconstrained generation is trivial, but generation under constraints can be NP-hard --- the real asymmetry is that parsing is \emph{always} constrained (the input is given) while generation need not be. Although the directional aspects of parsing (top-down vs.~bottom-up) and surprisal theory have been studied extensively (Russell, Carroll \& Warwick 1990; Hale 2001; Levy 2008), the explicit framing of directionality and temporality as distinct structural dimensions within a unified taxonomy of the generation-recognition asymmetry has not, to our knowledge, been articulated previously in formal language theory. We connect the temporal dimension to the surprisal framework of \citet{hale2001probabilistic} and \citet{levy2008expectation}, arguing that surprisal formalizes the temporal asymmetry between a generator that creates the future (surprisal = 0) and a parser that predicts under uncertainty (surprisal > 0).

We review bidirectional systems in NLP (DCG, Grammatical Framework, grammar inversion) and observe that bidirectionality has been available for over fifty years yet has not transferred to most domain-specific applications. We identify six counter-arguments to our framework and address each. We conclude with a discussion of large language models, which architecturally unify generation and recognition while operationally preserving the asymmetry.
\end{abstract}


\end{frontmatter}

\section{Introduction}

\subsection{The three uses of a grammar}

Every formal grammar \(G\) defines a language \(L(G)\) --- the set of all strings derivable from the start symbol by applying the production rules. This definition is direction-neutral: \(L(G)\) is simply a set. But the \emph{use} of \(G\) is not direction-neutral. A grammar can be employed in three fundamentally distinct ways:

\begin{enumerate}
\def\labelenumi{\arabic{enumi}.}
\item
  \textbf{Generation}: given \(G\), produce strings \(s \in L(G)\). The grammar is known; the output is sought. Starting from the axiom, apply rules to derive strings --- the \emph{modus ponens} of the grammar.
\item
  \textbf{Recognition} (parsing, analysis): given \(G\) and a string \(s\), determine whether \(s \in L(G)\) and, if so, assign it a structural description (parse tree). The grammar is known; the structure is sought. This is the \emph{abductive} counterpart --- observed effects \(\to\) probable causes: given the surface string, infer the derivation that produced it.
\item
  \textbf{Inference} (grammar induction): given a corpus of strings \(\{s_1, \ldots, s_k\}\), find a grammar \(G\) such that \(L(G)\) accounts for the observed data. Neither the grammar nor the structure is known; both are sought. This is a qualitatively different problem: recognition presupposes a known grammar, while inference must discover the grammar itself.
\end{enumerate}

\begin{longtable}[]{@{}
  >{\raggedright\arraybackslash}p{(\columnwidth - 6\tabcolsep) * \real{0.20}}
  >{\raggedright\arraybackslash}p{(\columnwidth - 6\tabcolsep) * \real{0.15}}
  >{\raggedright\arraybackslash}p{(\columnwidth - 6\tabcolsep) * \real{0.15}}
  >{\raggedright\arraybackslash}p{(\columnwidth - 6\tabcolsep) * \real{0.50}}@{}}
\toprule\noalign{}
\begin{minipage}[b]{\linewidth}\raggedright
Operation
\end{minipage} & \begin{minipage}[b]{\linewidth}\raggedright
Given
\end{minipage} & \begin{minipage}[b]{\linewidth}\raggedright
Sought
\end{minipage} & \begin{minipage}[b]{\linewidth}\raggedright
Analogy
\end{minipage} \\
\midrule\noalign{}
\endhead
\bottomrule\noalign{}
\endlastfoot
Generation & \(G\) & \(s\) & The speaker knows the rules and produces speech \\
Recognition & \(G + s\) & Structure & The listener knows the rules and parses speech \\
Inference & \(\{s_1, \ldots, s_k\}\) & \(G\) & The child has no rules and must learn the language \\
\end{longtable}

The first two uses pervade all applications of formal grammars. In linguistics, they correspond to speaking versus understanding. In compiler design, they correspond to code generation versus parsing. The third --- inference --- is the problem of language acquisition, of reverse-engineering a grammar from examples. It is widely regarded as the hardest of the three problems \citep[ch.~1]{delahiguera2010inference}, building on Gold's impossibility theorem \citep{gold1967language}, and the least studied in relation to the other two.

The relationship between generation and recognition is asymmetric in multiple, independent ways --- ways that have been studied individually but, to the best of our knowledge, never systematized. Inference adds a third axis that deepens the asymmetry: as we show in §4.5, generation, recognition, and inference form a hierarchy of increasing difficulty, not a simple duality.

Chomsky's original formulation was explicitly generative: a grammar is ``a device that generates all of the grammatical sequences [\ldots] and none of the ungrammatical ones'' (\citealt[p.~13]{chomsky1957syntactic}). The parsing problem --- given a string, determine its grammatical status and structural description --- emerged later as a distinct computational challenge, one that turned out to be substantially harder than generation for all but the most restricted formalisms. Grammar inference emerged later still, formalized by \citet{gold1967language} as ``identification in the limit.'' The \emph{Dragon Book} --- the standard reference in compiler theory --- devotes roughly two-thirds of its parsing-and-translation pipeline (chapters 4--5, \(\sim\)210 pages) to parsing strategies (LL, LR, Earley, CYK), compared with the \(\sim\)100 pages of chapter 8 on code generation, where the latter rests mainly on pattern matching and tree rewriting (\citealt{aho1986compilers}). Grammar inference receives no mention at all. This progressive disproportion --- generation taken for granted, recognition studied intensively, inference barely addressed --- is itself evidence of the asymmetry.

\subsection{A gap in the literature}

Although psycholinguistic frameworks have explicitly unified comprehension, production, and acquisition as related processes --- notably the \emph{P-chain} framework of \citet{dell2014pchain} and the survey of probabilistic models by \citet{chater2006probabilistic} --- \textbf{no prior work in formal language theory has, to our knowledge, proposed a comparable framework} that treats the asymmetry between generation, recognition, and inference as a multi-axis structural phenomenon characterizable by complexity classes, decidability bounds, and information-theoretic measures. The asymmetry is occasionally noted in passing in the formal-language literature --- \citet{russell1990asymmetry} on the practical asymmetry of unification grammars (title = ``Asymmetry in Parsing and Generating''); \citet{berwick1982complexity,berwick1984strong} on the computational-complexity gap between weak and strong generative capacity; \citet{strzalkowski1993reversible} on grammar-inversion techniques; \citet{shieber1988uniform} on the need to parameterize direction in a uniform processing architecture. But these observations remain scattered across separate literatures and do not propose a unified dimensional taxonomy.

In natural language processing, the distinction between NLG (Natural Language Generation) and NLU (Natural Language Understanding) is recognized as fundamental. \citet{karanikolas2023llm} survey the relationship between large language models and the NLG/NLU duality, and \citet{su2019dual} formalize NLU and NLG as a ``dual problem pair'' with shared latent variables. Yet no paper proposes a dimensional framework for analyzing the asymmetry itself --- the focus is on practical systems, not on the structure of the duality.

In formal language theory proper, the situation is similar: complexity results are known (CYK parsing is \(\mathcal{O}(n^3)\) for CFGs, while generation is \(\mathcal{O}(n)\) in derivation length), but these results are not connected to the determinism gap, the information gap, or the temporal gap that we identify below.

The \emph{P-chain} framework in psycholinguistics \citep{dell2014pchain,gambi2017models,martin2018prediction,gastaldon2024predictive} explicitly proposes that the production system implements implicit prediction during comprehension --- prediction error driving learning. Our framework operates at a different level of abstraction (formal language theory) and remains compatible with this psycholinguistic perspective while differing from it methodologically.

This paper fills that void in formal language theory. We propose a systematic framework of six dimensions along which generation and recognition diverge, drawing on evidence from compiler theory, natural language processing, information theory, and psycholinguistics.

\subsection{Against the naive view}

A common but misleading characterization of the asymmetry is ``generation is easy, analysis is hard.'' This is both too simple and partly wrong.

Unconstrained generation --- applying grammar rules at random until reaching terminal symbols --- is indeed trivial: \(\mathcal{O}(n)\) in derivation length for any formalism. The result is ``grammatical'' (it belongs to \(L(G)\)) but fundamentally useless. No one generates at random; real generation operates under constraints (a target meaning, a required form, a stylistic register). And generation under constraints can be as hard as, or harder than, parsing: \citet{song2016amr} formulate AMR-to-text generation as an asymmetric generalized traveling salesman problem (AGTSP); \citet{barton1987complexity} show that two-level morphology is NP-complete.

The real asymmetry is not a simple difficulty gap. It is \textbf{structural}: parsing is \emph{always} constrained --- the input string is given, non-negotiable --- while generation \emph{may or may not be}, depending on the task. When generation operates under tight constraints, it becomes as hard as parsing or harder. The asymmetry lies not in the intrinsic difficulty of either operation but in the fact that the recognizer has no choice: its difficulty is imposed by the input.

This observation has a corollary for modern AI systems. Large language models appear to generate effortlessly --- but they \textbf{displace} the analytical cost rather than eliminating it. The training of a language model IS a massive act of analysis: compressing a corpus into a parametric model. The generation phase (autoregressive sampling) is \(\mathcal{O}(n)\), but the cost has been paid upstream by analysis. \emph{Intelligent generation always presupposes prior analysis.}

\subsection{Contributions}

This paper makes seven contributions:

\begin{enumerate}
\def\labelenumi{\arabic{enumi}.}

\item
  A systematic identification and analysis of \textbf{six independent dimensions} of the generation-recognition asymmetry (§4).
\item
  Two dimensions --- \textbf{directionality} (D3, §4.3) and \textbf{temporality} (D6, §4.6) --- that, \textbf{to our knowledge}, have not been explicitly identified as structural dimensions within a unified taxonomy of this asymmetry in formal language theory, despite extensive study of their underlying aspects (Russell-Carroll-Warwick 1990; Hale 2001; Levy 2008; and the \emph{P-chain} framework in psycholinguistics, Dell \& Chang 2013).
\item
  A critique of the naive ``generation easy, parsing hard'' characterization: the real asymmetry is structural (the recognizer's difficulty is imposed, the generator's is chosen), not a simple difficulty gap (§1.3).
\item
  A formal-language-theory connection between the \textbf{surprisal framework} of \citet{hale2001probabilistic} and \citet{levy2008expectation} and the temporal dimension of the asymmetry (§4.6). This connection already exists in the psycholinguistic \emph{P-chain} perspective \citep{dell2014pchain}; our contribution is to situate it within the formal framework.
\item
  A review of \textbf{bidirectional systems} in NLP (§3.3) and the observation that bidirectionality has not transferred to most domain-specific formal systems despite being available since the 1970s.
\item
  Six \textbf{counter-arguments} to our framework, identified and addressed (§6.1, Appendix A).
\item
  A discussion of \textbf{large language models} as an apparent counter-example to the asymmetry thesis, and why the asymmetry persists in displaced form (§6.5).
\end{enumerate}

\subsection{Scope and non-scope}

This paper is domain-independent: it addresses the asymmetry at the level of formal language theory, with examples drawn from compiler theory, natural language processing, and information theory. Domain-specific manifestations --- in particular, the rich case of music, where both directions exist naturally (composition vs.~analysis) --- are treated in a companion paper (Author, in preparation). We present no experimental results; this is a critical survey with an original thesis.

\section{Theoretical Framework}

The generation-recognition asymmetry can be grounded in three classical frameworks, each illuminating a different facet: Shannon's communication model (the information-theoretic dimension), the Chomsky hierarchy (the complexity dimension), and Morris's semiotic triangle (the semantic dimension). Together, they provide the theoretical scaffolding for the six dimensions analyzed in §4.

\subsection{Shannon: the encoder-decoder asymmetry}

Shannon's mathematical theory of communication \citeyearpar{shannon1948mathematical} models information transmission as:

\begin{figure}[H]
\centering
\resizebox{\columnwidth}{!}{%
% Figure — Shannon's Communication Model and the Generation-Recognition Analogy
\begin{tikzpicture}[scale=1.2, transform shape, every node/.style={font=\small},
  box/.style={draw=black!50, rounded corners, fill=white, text=black, font=\normalsize\bfseries, minimum width=1.8cm, minimum height=0.8cm}]
  % Shannon model (top row)
  \node[font=\footnotesize\itshape, text=black!50] at (-1.5,2.5) {Shannon:};
  \node[box] (src) at (0,2.5) {Source};
  \node[box, draw=blue!50, fill=blue!8] (enc) at (3,2.5) {Encoder};
  \node[box, draw=black!40, fill=black!5] (ch) at (6,2.5) {Channel};
  \node[box, draw=red!50, fill=red!8] (dec) at (9,2.5) {Decoder};
  \node[box] (dst) at (12,2.5) {Destination};
  \draw[->, thick, black!50] (src) -- (enc);
  \draw[->, thick, black!50] (enc) -- (ch) node[midway, above, font=\footnotesize, text=black!60] {\(x\)};
  \draw[->, thick, black!50] (ch) -- (dec) node[midway, above, font=\footnotesize, text=black!60] {\(y\)};
  \draw[->, thick, black!50] (dec) -- (dst);
  \node[above=0.1cm, font=\footnotesize, text=black!60] at (ch.north) {\(+ \text{noise}\)};
  % Grammar analogy (bottom row)
  \node[font=\footnotesize\itshape, text=black!50] at (-1.5,0.8) {Grammar:};
  \node[box] (int) at (0,0.8) {Intention};
  \node[box, draw=blue!50, fill=blue!8] (gen) at (3,0.8) {Generator};
  \node[box, draw=black!40, fill=black!5] (lin) at (6,0.8) {Linearization};
  \node[box, draw=red!50, fill=red!8] (par) at (9,0.8) {Parser};
  \node[box] (str) at (12,0.8) {Structure};
  \draw[->, thick, black!50] (int) -- (gen);
  \draw[->, thick, black!50] (gen) -- (lin) node[midway, above, font=\footnotesize, text=black!60] {tree};
  \draw[->, thick, black!50] (lin) -- (par) node[midway, above, font=\footnotesize, text=black!60] {string};
  \draw[->, thick, black!50] (par) -- (str);
  \node[above=0.1cm, font=\footnotesize, text=black!60] at (lin.north) {info loss};
  % Annotations
  \node[text=blue!70!black, font=\footnotesize\bfseries] at (3,1.6) {\(H(X|X)=0\)};
  \node[text=red!70!black, font=\footnotesize\bfseries] at (9,1.6) {\(H(X|Y)>0\)};
\end{tikzpicture}
}
\caption{Shannon's communication model and the generation-recognition analogy. The encoder (generator) knows the source with certainty; the decoder (recognizer) must infer it under equivocation $H(X|Y) > 0$.}
\label{fig:shannon}
\end{figure}

The encoder transforms a message \(x\) into a signal; the decoder reconstructs the message from a possibly corrupted signal \(y\). The two agents are \textbf{structurally asymmetric}: the encoder knows \(x\) with certainty (it created it), while the decoder must infer \(x\) from \(y\), suffering the equivocation \(H(X|Y) > 0\) introduced by the noisy channel.

The mutual information \(I(X; Y) = H(X) - H(X|Y)\) quantifies how much of the original message survives the channel. The encoder's task is bounded by the source entropy \(H(X)\) --- it must represent the message efficiently. The decoder's task is bounded by the channel capacity \(C = \max I(X; Y)\) --- it must recover the message despite channel degradation. These are fundamentally different problems: the encoder faces a \emph{representation} problem (compressing the source to its entropy limit), while the decoder faces a \emph{reconstruction} problem (inferring the original message from a corrupted signal).

The analogy with generation and recognition is direct. The generator is the encoder: it transforms an intention (the source) into a string (the signal), choosing among the derivations licensed by the grammar. The recognizer is the decoder: it receives the string and must reconstruct its structural description --- the derivation that produced it --- from the surface alone. The ``noise'' in this analogy is the \textbf{information loss} inherent in linearization: a hierarchical structure (the parse tree) is flattened into a sequence of symbols, and the recognizer must recover the hierarchy from the sequence. This is a non-trivial inversion.

\citet{eco1979lector}, building on \citet{jakobson1960linguistics}, adds a semiotic nuance: the text is ``a lazy mechanism'' (\emph{meccanismo pigro}) that ``lives on the surplus of meaning introduced by the addressee.'' The generator deliberately omits information (through what Eco calls the ``narcotization of properties''), trusting the decoder to fill in the gaps. This is not a defect of communication but a feature: the asymmetry is \textbf{designed into} expressive systems.

\subsection{Chomsky: the hierarchy as complexity ladder}

The Chomsky hierarchy classifies formal grammars by their generative power:

\begin{longtable}[]{@{}cll@{}}
\toprule\noalign{}
Type & Class & Automaton \\
\midrule\noalign{}
\endhead
\bottomrule\noalign{}
\endlastfoot
3 & Regular & Finite automaton \\
2 & Context-free & Pushdown automaton \\
2+ & Mildly context-sensitive (TAG, CCG, MCFG) & Embedded pushdown \\
1 & Context-sensitive & Linear-bounded automaton \\
0 & Recursively enumerable & Turing machine \\
\end{longtable}

Each level adds expressive power but at a cost that falls \textbf{asymmetrically} on generation and recognition. For terminating grammars (Types 1--3), unconstrained generation is \(\mathcal{O}(n)\) at all levels, while recognition grows from linear to cubic, polynomial, exponential, and ultimately undecidable. The detailed complexity analysis is developed in §4.1 (Table 4).

\citet{miller2000strong} distinguishes weak generative capacity (the set of strings generated) from strong generative capacity (the set of structural descriptions assigned). We observe that this distinction has an asymmetric impact: the generator need only produce \emph{a} valid string --- weak capacity suffices. The recognizer, however, must determine \emph{which structure(s)} the string admits --- strong capacity is what counts. Two grammars may generate the same strings yet assign different structures; this difference is invisible from the generation side but critical from the recognition side.

\subsection{Morris: the semiotic triangle}

\citet{morris1938foundations} partitioned semiotics into three domains:

\begin{itemize}

\item
  \textbf{Syntactics}: the formal relations among signs (grammar, derivation rules)
\item
  \textbf{Semantics}: the relation between signs and what they denote (meaning, interpretation)
\item
  \textbf{Pragmatics}: the relation between signs and their users (context, intention, use)
\end{itemize}

The generation-recognition asymmetry cuts across all three. At the syntactic level, the generator applies rules (top-down derivation), while the recognizer inverts them (parsing). At the semantic level, the generator maps from meaning to form (the speaker knows what they want to say), while the recognizer maps from form to meaning (the listener must infer intent from surface). At the pragmatic level, the generator acts within a context they control (choosing what to say and how), while the recognizer must reconstruct that context from indirect evidence.

The commutative diagram of translation --- drawn from the categorial tradition in mathematical logic (\citealt{lambek1958mathematics}; \citealt{lawvere1963functorial}) --- illustrates this. Given two languages \(L_1\) and \(L_2\) with semantics functions \(\mathcal{S}_1\) and \(\mathcal{S}_2\), a translation \(T: L_1 \to L_2\) preserves meaning if the following diagram commutes:

\begin{figure}[H]
\centering
\resizebox{0.5\columnwidth}{!}{%
% Figure 0 — Commutative Diagram: Round-Trip Test
\begin{tikzpicture}[scale=1.5, transform shape, every node/.style={font=\normalsize, text=black}]
  \node (L1) at (0,1.5) {\(L_1\) (Form\(_{1}\))};
  \node (L2) at (3,1.5) {\(L_2\) (Form\(_{2}\))};
  \node (M1) at (0,0) {\(\mathcal{M}_1\) (Meaning\(_{1}\))};
  \node (M2) at (3,0) {\(\mathcal{M}_2\) (Meaning\(_{2}\))};
  \draw[->, thick, black!50] (L1) -- (L2) node[midway, above, text=black] {\(T\)};
  \draw[->, thick, black!50] (L1) -- (M1) node[midway, left, text=black] {\(\mathcal{S}_1\)};
  \draw[->, thick, black!50] (L2) -- (M2) node[midway, right, text=black] {\(\mathcal{S}_2\)};
  \draw[double, thick, black!50] (M1) -- (M2) node[midway, below, text=black] {\(=\)};
\end{tikzpicture}
}
\caption{Commutative diagram: the round-trip test. Generation followed by recognition should recover the original structure — the failure of this commutativity measures the asymmetry.}
\label{fig:commutative}
\end{figure}

That is, \(\mathcal{S}_2(T(s)) = \mathcal{S}_1(s)\): translating and then interpreting yields the same meaning as interpreting directly. When the ``translation'' is the round-trip generation → recognition --- encoding an intention into a string and then decoding it back --- this commutativity condition becomes a \textbf{round-trip test}: does the recognizer recover the structure that the generator intended? The failure of this test, for any non-trivial grammar, is a measure of the asymmetry.

\subsection{Synthesis: three frameworks, one asymmetry}

The three frameworks converge on a single structural observation. Shannon shows that the encoder (generator) and decoder (recognizer) face fundamentally different problems: the encoder knows the message, the decoder must infer it from a noisy signal. Chomsky shows that this asymmetry grows with the expressive power of the formalism: generation remains \(\mathcal{O}(n)\) while recognition escalates from \(\mathcal{O}(n)\) to undecidable. Morris shows that the asymmetry cuts across all semiotic levels --- syntactic, semantic, and pragmatic --- and that the round-trip test (generation followed by recognition) is a measure of how much information is lost in linearization.

Together, these frameworks establish that the generation-recognition asymmetry is not an accident of particular algorithms or implementations but a \textbf{structural property of the relationship between hierarchical structure and its linear projection} --- it arises wherever multi-dimensional information must pass through a sequential channel. Formal systems do not create this asymmetry; they reveal it and make it quantifiable. The six dimensions analyzed in §4 are concrete manifestations of this structural asymmetry, each irreducible to the others.

\section{The Landscape}

Having established the theoretical framework, we survey the principal approaches, algorithms, and formalisms that constitute the generation-recognition landscape. We organize the survey by direction --- analytical, generative, and bidirectional --- then propose a taxonomy. For each direction, we adopt a comparative lens: what does each approach reveal not only about the direction it serves, but also about the direction it does not? This systematic cross-reading prepares the formal analysis of §4.

\subsection{Analytical systems}

Parsing --- the recognition direction --- has been the dominant concern of compiler theory and computational linguistics for over sixty years. The richness and variety of parsing algorithms far exceeds that of generation algorithms, a disproportion that is itself evidence of the asymmetry.

\textbf{In compiler design}, parsing is the core of the front end. The major paradigms --- LL (top-down predictive), LR (bottom-up shift-reduce), Earley (mixed top-down/bottom-up), CYK (bottom-up dynamic programming), GLR (generalized LR) --- represent decades of sustained research to handle increasingly expressive grammars efficiently. Knuth's LR parsing \citeyearpar{knuth1965translation} and Earley's algorithm \citeyearpar{earley1970efficient} remain foundational. By contrast, code generation --- the compiler's generative phase --- received comparatively less algorithmic attention, relying primarily on pattern matching and tree rewriting (\citealt{aho1986compilers}, Part III).

\textbf{In natural language understanding}, parsing has progressed through multiple paradigms: phrase-structure parsing (Collins, 1997; Charniak, 2000), dependency parsing (\citealt{nivre2008algorithms}; Chen \& Manning, 2014), semantic parsing (Zettlemoyer \& Collins, 2005), and discourse parsing (Joshi \& Schabes, 1997). Each paradigm addresses a different level of linguistic structure, but all share the same direction: given a string, assign it a structure. The progressive deepening --- from syntactic to semantic to discourse --- has no parallel in generation, where ``deep generation'' (from communicative intentions) has remained a comparatively niche endeavor.

\textbf{Constraint-based approaches} --- unification grammars (HPSG, LFG), constraint satisfaction for parsing --- further illustrate the analytical emphasis. The unification operation is fundamentally an analytical tool: it takes partial structures and combines them, resolving ambiguities through constraint propagation. While unification grammars are in principle declarative and direction-neutral, their practical implementations are overwhelmingly analytical (Shieber, 1988; \citealt{wintner1997amalia}).

\subsection{Generative systems}

The generative direction --- given a grammar, produce strings --- has received less systematic attention in the formal language theory literature, though it is the bread and butter of practical applications (template-based NLG, code generation, procedural content generation).

\textbf{The NLG pipeline} of \citet{reiter2000building} decomposes generation into three stages: Document Planning (deciding \emph{what} to say), Microplanning (deciding \emph{how} to say it: lexical choice, aggregation, referring expressions), and Surface Realization (producing the syntactic string). This pipeline is structurally top-down: high-level intentions are progressively refined into surface forms. The parsing pipeline runs in the opposite direction, from surface to structure.

\textbf{Template-based generation} represents the simplest generative paradigm: predefined patterns with slots. It is efficient (linear time) but rigid --- templates cannot generalize beyond their design. The popularity of template-based systems in practical NLG (weather reports, sports summaries, medical reports) testifies to the fact that \emph{constrained} generation is the common case: the generator must satisfy specifications, not produce arbitrary strings. There is no analytical counterpart to template-based generation: recognizing a string's structure cannot be reduced to slot-matching, because the string must be decomposed into its hierarchical constituents --- a fundamentally harder operation that requires the full apparatus of parsing.

\textbf{Neural generation} (language models, sequence-to-sequence architectures, diffusion models) has transformed NLG since 2018. Large language models generate fluent text by autoregressive sampling --- a process that is \(\mathcal{O}(n)\) per token, superficially resembling unconstrained generation. But the cost has been displaced, not eliminated: the training phase is an act of massive analysis (compressing a corpus into model parameters), and the generation quality depends entirely on the quality of this prior analysis. Adding constraints at generation time (format, factuality, safety) further increases difficulty, potentially to NP-hardness. The asymmetry takes a modern form: autoregressive generation is fast (\(\mathcal{O}(n)\) per token), but the training that enables it is computationally enormous --- billions of parameters learned from trillions of tokens. Generation is cheap \emph{because} analysis was expensive.

\textbf{A recurring observation}: the difficulty of generation lies not in producing \emph{some} string but in producing the \emph{right} string, one that satisfies pragmatic, semantic, or aesthetic constraints. Unconstrained generation is trivially easy for any grammar --- one applies rules at random until reaching terminal symbols. The result is ``grammatical'' but useless. This is why the NLG literature focuses so heavily on planning and realization: the hard problem is not derivation but constraint satisfaction.

\subsection{Bidirectional systems: the evidence from NLP}

A small but significant body of work in computational linguistics has pursued the goal of using the \textbf{same grammar} for both generation and parsing. This literature provides direct evidence about the nature of the asymmetry --- and about the conditions under which it can be bridged.

\textbf{Definite Clause Grammars (DCG)}, embedded in Prolog, are inherently bidirectional: the same grammar can be queried in either direction, using Prolog's unification and backtracking. However, this symmetry holds at the level of the grammar notation, not at the level of execution. Left-recursive rules cause infinite loops in top-down parsing but not in generation; conversely, right-recursive rules can cause generation to diverge. The same grammar, used in two directions, encounters \emph{different} pathological cases --- a concrete manifestation of the directionality asymmetry (D3).

\textbf{Finite-state transducers (FST)} in computational morphology achieve bidirectionality through a clean mathematical property: the inverse of an FST is also an FST. If \(T\) maps underlying forms to surface forms (generation), then \(T^{-1}\) maps surface forms to underlying forms (analysis). Koskenniemi's two-level morphology (1983) exploits this directly. But the inversion property holds only for finite-state machines --- it does not generalize to context-free or context-sensitive grammars, where inversion is undecidable in general \citep{strzalkowski1993reversible}.

\textbf{Q-systems} (\citealt{colmerauer1970systemes}), the precursor to Prolog, were designed from the outset for reversible computation. Their influence persists in the logic programming tradition, where the separation of logic from control (\citealt{kowalski1979algorithm}) makes direction a parameter rather than a fixed architectural choice.

\textbf{Grammatical Framework (GF)} (\citealt{ranta2004grammatical,ranta2019grammatical}) is perhaps the most developed modern bidirectional grammar formalism. GF separates abstract syntax (language-independent meaning representation) from concrete syntax (language-specific surface realization). The same abstract syntax tree can be linearized (generation) or parsed (recognition) in any of the supported languages. The Portable Grammar Format (PGF) provides efficient runtime support for both directions. GF demonstrates that bidirectionality is achievable at scale --- the GF Resource Grammar Library covers over 40 languages (\citealt{ranta2009resource}) --- but requires a specific architectural commitment: the grammar must be written declaratively, with meaning and form cleanly separated.

\textbf{Appelt's KAMP system} (1985, 1987) attempted a unified architecture for planning and realization in NLG. Appelt argued that ``the most fundamental requirement of any bidirectional grammar is that it be represented \textbf{declaratively}. If procedural, asymmetry is inevitable'' (\citealt{appelt1987bidirectional}). However, KAMP was described as ``computationally impractical'' --- the cost of unifying planning and realization in a single framework proved prohibitive. This is a cautionary result: bidirectionality is possible in principle but expensive in practice.

\textbf{\citet{strzalkowski1993reversible,strzalkowski1994general}} pursued a different approach: \textbf{grammar inversion}. Rather than designing a grammar to be bidirectional from the start, Strzalkowski developed techniques to automatically transform a parsing grammar into a generation grammar (and vice versa) by inverting the control structure while preserving the declarative content. The results showed that inversion is feasible but non-trivial: the inverted grammar may be less efficient, and certain constructions resist inversion entirely.

\textbf{\citet{goodman2009generation}} provided empirical evidence for the value of bidirectionality: using a grammar for both parsing and generation reveals errors that would remain hidden in unidirectional use. Their experiments with a broad-coverage HPSG grammar showed an 18\% increase in coverage when the grammar was tested bidirectionally, because generation exposed semantic gaps that parsing alone could not detect.

\textbf{A striking observation emerges from this survey}: bidirectional grammar systems have been available in NLP since the 1970s (DCG, FST, Q-systems) and have been refined over five decades (GF, Amalia, grammar inversion). Yet this technology has \textbf{not transferred} to most domain-specific applications of formal grammars --- in bioinformatics (where sequence grammars are used for RNA structure prediction but not generation), in computer-aided design (where shape grammars generate but rarely analyze), or in computational musicology (where grammars generate or analyze but almost never both). Why?

We propose two hypotheses. First, bidirectionality requires \textbf{declarativity} (\citealt{appelt1987bidirectional}): the grammar must separate what it says from how it is processed. Most domain-specific formalisms are procedural --- they embed the processing direction in the grammar itself --- making inversion difficult or impossible. Second, the \textbf{cost of bidirectionality is hidden}: systems like KAMP that attempt full integration are ``computationally impractical,'' and the perceived benefit (bidirectional error detection) does not justify the engineering effort in domains where only one direction is needed. These hypotheses are speculative; we offer them as directions for future work.

\subsection{A taxonomy}

The following table organizes the systems surveyed above by their supported directions:

\begin{table}[H]
\centering\small
\begin{tabular}{@{}
  >{\raggedright\arraybackslash}p{(\columnwidth - 10\tabcolsep) * \real{0.43}}
  >{\raggedright\arraybackslash}p{(\columnwidth - 10\tabcolsep) * \real{0.11}}
  >{\centering\arraybackslash}p{(\columnwidth - 10\tabcolsep) * \real{0.11}}
  >{\centering\arraybackslash}p{(\columnwidth - 10\tabcolsep) * \real{0.12}}
  >{\centering\arraybackslash}p{(\columnwidth - 10\tabcolsep) * \real{0.12}}
  >{\centering\arraybackslash}p{(\columnwidth - 10\tabcolsep) * \real{0.11}}@{}}

\toprule\noalign{}
\begin{minipage}[b]{\linewidth}\raggedright
System
\end{minipage} & \begin{minipage}[b]{\linewidth}\raggedright
Domain
\end{minipage} & \begin{minipage}[b]{\linewidth}\centering
Generation
\end{minipage} & \begin{minipage}[b]{\linewidth}\centering
Recognition
\end{minipage} & \begin{minipage}[b]{\linewidth}\centering
Inference
\end{minipage} & \begin{minipage}[b]{\linewidth}\centering
Bidirectional
\end{minipage} \\
\midrule\noalign{}
\bottomrule\noalign{}
LL/LR parsers & Compilers & --- & $\checkmark$ & --- & --- \\
Earley/CYK & FL* & --- & $\checkmark$ & --- & --- \\
NLG pipeline (Reiter \& Dale) & NLP & $\checkmark$ & --- & --- & --- \\
Template-based NLG & NLP & $\checkmark$ & --- & --- & --- \\
LLMs (GPT, etc.) & NLP & $\checkmark$ & (implicit) & --- & --- \\
DCG (Prolog) & NLP & $\checkmark$ & $\checkmark$ & --- & $\checkmark$* \\
FST (Koskenniemi) & Morphology & $\checkmark$ & $\checkmark$ & --- & $\checkmark$ \\
GF (Ranta) & NLP & $\checkmark$ & $\checkmark$ & --- & $\checkmark$ \\
KAMP (Appelt) & NLP & $\checkmark$ & $\checkmark$ & --- & $\checkmark$* \\
Grammar inversion (Strzalkowski) & NLP & $\checkmark$ & $\checkmark$ & --- & $\checkmark$ \\
LSTAR (Angluin) & FL** & --- & --- & $\checkmark$ & --- \\
Gold's framework & FL** & --- & --- & $\checkmark$ & --- \\
\end{tabular}

\vspace{2pt}\noindent{\footnotesize *Bidirectional in principle, with asymmetric pathologies or impractical cost. **FL = Formal languages.}
\end{table}

This taxonomy is illustrative, not exhaustive --- it samples representative systems from the survey above. Three qualitative observations emerge. First, \textbf{bidirectional systems cluster in NLP}, where the practical need for both directions (machine translation, dialogue systems, grammar engineering) has driven half a century of development; in other domains (compiler design, bioinformatics, computational musicology), unidirectional systems remain the norm. Second, even bidirectional systems exhibit \textbf{residual asymmetries} --- DCG's direction-dependent pathologies, KAMP's prohibitive cost --- suggesting that the asymmetry cannot be fully engineered away. Third, \textbf{grammar inference} stands apart as a ``third direction'' that is rarely combined with either generation or recognition in a single system --- a separation we analyze as Dimension 5 (§4.5).

\section{Six Dimensions of the Asymmetry}

This section identifies and analyzes six independent dimensions along which generation and recognition diverge. For each dimension, we state a thesis, provide a formal argument, illustrate with examples, and address potential counter-arguments. The dimensions are summarized in Appendix~A.

\subsection{D1 --- Computational Asymmetry}

\textbf{Thesis.} The computational cost of recognizing a string in the language defined by a grammar grows with the position of that grammar in the Chomsky hierarchy, whereas generating an arbitrary string of the same language remains polynomial at every level where the derivation terminates. The gap between the two operations is therefore not uniform. It is null for regular languages, where both tasks run in linear time. It is polynomial for CFGs (of the order of \(n^2\) to \(n^4\) depending on the formalism chosen, and of the order of \(n^3\) to \(n^4\) for TAG/MCFG). It separates two conjecturally distinct complexity classes for CSGs, where recognition is PSPACE-complete while generation remains polynomial. It finally reaches the threshold of decidability for recursively enumerable languages, where recognition is no longer computable in general.

\textbf{Methodological framework.} All the bounds that follow count the \emph{elementary operations} executed by a deterministic Turing machine on the input string, in the worst case, for a grammar of fixed size (the \(|G|\) factors are mentioned explicitly when they intervene). We distinguish two types of bounds. An \emph{intrinsic bound} is a \textbf{tight} characterization of the problem's complexity: when a single function \(f\) serves as both an upper and a lower bound (up to a constant factor), we speak of a tight bound and write it \(\Theta(f(n))\), defined as the intersection \(O(f(n)) \cap \Omega(f(n))\). Alternatively, the characterization may take the form of a complexity class (PSPACE-complete, undecidable). An intrinsic bound cannot be improved algorithmically, short of proving an equality between two conjecturally distinct complexity classes (for example \(P = PSPACE\), a result widely presumed false). A \emph{best-known-algorithm bound} is the tightest upper bound that a currently published algorithm attains for the problem; it may in principle be improved by a future algorithm.

\textbf{Six sub-problems to separate.} The generic terms ``generation'' and ``recognition'' in fact each cover three distinct computational sub-problems, which must be stated before any bound is computed. \emph{On the recognition side}, we distinguish (i) the \emph{membership decision} --- determining whether a given string \(w\) belongs to the language \(L(G)\), that is, answering yes or no; (ii) the \emph{construction of a derivation tree} --- producing one valid structural description of \(w\), a problem whose bound differs from the membership decision only by a constant factor; and (iii) the \emph{enumeration of all trees} --- listing all valid derivation trees for \(w\), a problem whose complexity can explode combinatorially, up to \(\Theta(C_n) \approx 4^n / n^{3/2}\) distinct trees for an ambiguous CFG string (where \(C_n\) denotes the \(n\)-th Catalan number). \emph{On the generation side}, we distinguish three parallel sub-problems: (i) \emph{free generation} --- applying the grammar rules with no particular target and stopping at an arbitrary step, the cost being counted in derivation steps; (ii) \emph{terminating example-generation} --- producing one complete string of \(L(G)\) from the axiom (a sentential form with no non-terminals), the cost being counted in total elementary operations, rule applications included; and (iii) \emph{generation under semantic constraint} --- producing a string \(w\) that additionally satisfies an external constraint \(C\) (a target meaning, a pragmatic requirement, an imposed format), an NP-complete problem in general even for CFGs \citep{barton1987complexity}. These six sub-problems are examined in turn in the two subsections below (recognition side, then generation side), before being synthesized in two compared heatmaps that bring out the differential coupling central to D1.

\subsubsection*{Recognition side}

\textbf{Membership decision.} This first bound --- around which the entire analysis below is organized --- determines whether or not a given string belongs to the language defined by the grammar. It depends strongly on the class in the hierarchy.

For a \textbf{regular} language (\(k=3\)), the membership decision is solved in linear time \(\Theta(n)\) as soon as a pre-compiled \textbf{deterministic finite automaton} (DFA) is available. A DFA is defined by a finite set of states \(Q\), an alphabet \(\Sigma\), an initial state \(q_0 \in Q\), a subset of final states \(F \subseteq Q\), and a transition function \(\delta : Q \times \Sigma \to Q\) that maps each pair (current state, symbol read) to \textbf{one and only one} next state --- it is this uniqueness that characterizes determinism, as opposed to the nondeterministic finite automaton (NFA) introduced later, where \(\delta\) may return a set of successor states. The bound is tight for two independent reasons: on the one hand, direct simulation of the DFA gives the upper bound \(O(n)\), since each input character triggers exactly one transition; on the other hand, the lower bound \(\Omega(n)\) is trivial, since the entire string must necessarily be read \citep[ch.~2--4]{hopcroft2006introduction}.

This bound assumes, however, that the DFA is known in advance. If the grammar is supplied in a less direct form --- typically a regular expression (regex) or an equivalent nondeterministic automaton (NFA), formats in which regular languages are naturally written --- one must first construct an equivalent DFA by \emph{determinization}. This construction can produce up to \(O(2^{|Q|})\) states in the worst case, where \(|Q|\) denotes the number of states of the initial NFA; performed on the fly while reading the string, it brings the total cost to \(O(n \cdot |Q|^2)\).

For a \textbf{context-free} grammar (\(k=2\)), the membership decision is in the class \(\mathsf{P}\), but its exact intrinsic complexity remains an open problem. The classical algorithm of \citet{younger1967recognition}, now known as CYK, attains \(O(n^3 \cdot |G|)\) for a grammar in \textbf{Chomsky normal form} --- that is, a grammar whose production rules all take one of two forms: either \(A \to BC\) (a non-terminal into two non-terminals), or \(A \to a\) (a non-terminal into a terminal). Every CFG admits an equivalent CFG in this normal form (with no loss of expressive power), and CYK requires this format because it recursively decomposes each input substring into two parts:

\begin{quote}
\emph{``A recognition algorithm is exhibited [\ldots] completed in a number of steps proportional to the `cube' of the number of symbols in the tested string.''} \citep[p.~189]{younger1967recognition}
\end{quote}

\citet{valiant1975general} brought this bound below cubic by reducing CFG recognition to Boolean matrix multiplication (BMM hereafter):

\begin{quote}
\emph{``context-free recognition, for \(n\) character input strings, can be carried out at least as fast as multiplication for \(n \times n\) Boolean matrices. Using Strassen's method [\ldots] an indirect algorithm for general context-free recognition can be derived that has time complexity \(O(n^{2.81})\).''} \citep[p.~308]{valiant1975general}
\end{quote}

According to the best bound known at the time of writing for the matrix-multiplication exponent, \(\omega \leq 2.371552\) proved by \citet{williams2024new} (improving the result \(\omega < 2.3728639\) of \citealt{legall2014powers}), the bound becomes \(O(n^\omega) \approx O(n^{2.37})\). This value of \(\omega\) is liable to be improved by later work; any such improvement will carry over mechanically to the CFG-parsing bound. \citet{lee2002fast} established the converse: any algorithm that parsed CFGs in time below \(O(n^{3-\epsilon})\) would, by the same token, yield a Boolean matrix-multiplication algorithm below \(O(m^{3-\epsilon/3})\). The intrinsic complexity of the membership decision for CFGs is therefore \emph{conditionally equivalent} to that of Boolean matrix multiplication, with no unconditional tight characterization known to date.

For \textbf{mildly context-sensitive} grammars (\(k=2^+\)) --- including Tree-Adjoining Grammars (TAG), fixed-rank Multiple Context-Free Grammars (MCFG), Combinatory Categorial Grammars (CCG), and Linear Context-Free Rewriting Systems (LCFRS), all equivalent in weak expressive power according to \citet{vijayshankerweir1994equivalence} --- the membership decision is in \(\mathsf{P}\). The best known bound for TAG parsing is \(O(|G| \cdot n^6)\), obtained by tabular methods \citep[§10.5]{joshischabes1997tree}. \citet{satta1994tag} established that it is conditionally tight:

\begin{quote}
\emph{``any algorithm for TAG parsing that improves the \(O(|G||w|^6)\) time upper bound can be converted into an algorithm for Boolean matrix multiplication running in less than \(O(m^3)\) time.''} \citep[p.~174]{satta1994tag}
\end{quote}

For rank-\(m\) MCFGs, the bound grows polynomially with \(m\) (typically of the order of \(n^{2m+2}\) depending on the definition adopted).

For \textbf{context-sensitive} grammars (\(k=1\)), the membership decision is \textbf{PSPACE-complete}, where \(\mathsf{PSPACE}\) denotes the class of decision problems solvable by a deterministic Turing machine in \emph{polynomial space} (that is, using at most \(n^k\) tape cells for some constant \(k\) and an input of length \(n\)); a problem is \emph{PSPACE-complete} if it belongs to this class and represents the maximal difficulty found within it, in the sense that every other problem in the class reduces to it in polynomial time. This is an intrinsic tight characterization, whose establishment combines three steps. \citet{kuroda1964classes} establishes the equivalence between CSLs and the languages recognized by nondeterministic linear-bounded automata (NLBA). \citet{savitch1970relationships} then proves the inclusion \(\mathsf{NSPACE}(L(n)) \subseteq \mathsf{DSPACE}(L(n)^2)\) for every function \(L(n) \geq \log_2 n\), where \(\mathsf{DSPACE}(f)\) and \(\mathsf{NSPACE}(f)\) denote respectively the classes of languages decidable in space \(f(n)\) by a \textbf{deterministic} and a \textbf{nondeterministic} Turing machine. Combined with Kuroda, this inclusion places the CSLs in \(\mathsf{DSPACE}(n^2)\), and hence \emph{a fortiori} in \(\mathsf{PSPACE} = \bigcup_k \mathsf{DSPACE}(n^k)\):

\begin{quote}
\emph{``every context-sensitive language can be recognized within deterministic storage \(n^2\), where \(n\) is the length of the input.''} \citep[p.~177]{savitch1970relationships}
\end{quote}

PSPACE-hardness is then established in the classical way \citep[Thm~11.3]{hopcroft2006introduction}, following the work of \citet{stockmeyermeyer1973word} on word problems requiring exponential space. No polynomial-\emph{time} algorithm for the membership decision in CSLs is known, and there probably is none: the intrinsic bound would collapse only on condition of proving \(P = PSPACE\), which is widely presumed false. The best known \emph{time} bound is therefore exponential, more precisely \(O(2^{O(n^2)})\) --- obtained by enumerating the configurations reachable by the machine in space \(n^2\), whose total number is bounded by \(2^{O(n^2)}\) --- for a \emph{space} bound of \(O(n^2)\) according to Savitch.

For \textbf{recursively enumerable} languages (\(k=0\)), finally, the membership decision is \textbf{undecidable} \citep{turing1936computable} --- an intrinsic tight characterization, equivalent to the halting problem. Membership is only \emph{semi-decidable}: if \(w \in L(G)\), a naive enumeration halts in finite time; if \(w \notin L(G)\), it may never halt.

\begin{table}[H]
\centering
\small
\caption{Membership-decision bounds by class of the Chomsky hierarchy.}
\label{tab:d1-membership}
\begin{tabular}{@{}l p{0.40\columnwidth} p{0.34\columnwidth}@{}}
\toprule
Class & Best algorithmic bound & Intrinsic characterization \\
\midrule
\(k=3\) Regular & \(\Theta(n)\) by DFA (Hopcroft et al.\ 2006) & tight bound \(\Theta(n)\) \\
\(k=2\) CFG & \(O(n^3)\) by CYK (Younger 1967); \(O(n^\omega)\) with \(\omega \leq 2.37\) (Valiant 1975; Williams et al.\ 2024) & in \(\mathsf{P}\), conditionally equivalent to BMM (Lee 2002) \\
\(k=2^+\) TAG/MCFG & \(O(n^6)\) (Joshi \& Schabes 1997) & in \(\mathsf{P}\), conditionally tight, BMM reduction (Satta 1994) \\
\(k=1\) CSG & \(O(2^{n^2})\) in time, \(O(n^2)\) in space (Savitch 1970) & \textbf{PSPACE-complete} (Kuroda 1964; Savitch 1970; Stockmeyer \& Meyer 1973) \\
\(k=0\) RE & not applicable & \textbf{undecidable} (Turing 1936) \\
\bottomrule
\end{tabular}
\end{table}

\begin{figure}[H]
\centering
\resizebox{\columnwidth}{!}{%
\begin{tikzpicture}[scale=1.3, transform shape]
\def\xmax{13}
\def\ymax{8.5}
\def\xf{0.3714}
\def\yf{0.7727}
\foreach \i in {0,1,...,11} {
  \draw[black!12] (0,\i*\yf) -- (\xmax,\i*\yf);
}
\foreach \i in {5,10,15,20,25,30,35} {
  \draw[black!12] (\i*\xf,0) -- (\i*\xf,\ymax);
}
\draw[black!60, line width=0.6pt, -{latex}] (0,0) -- (\xmax+0.4,0);
\draw[black!60, line width=0.6pt, -{latex}] (0,0) -- (0,\ymax+0.4);
\foreach \v in {5,10,15,20,25,30,35} {
  \draw[black!50] (\v*\xf, -0.08) -- (\v*\xf, 0.08);
  \node[font=\scriptsize, text=black, below] at (\v*\xf, -0.1) {\v};
}
\node[font=\footnotesize, text=black] at (\xmax/2, -0.7) {\(n\) (input length)};
\foreach \i/\lab in {0/\(1\), 1/\(10\), 2/\(10^2\), 3/\(10^3\), 4/\(10^4\), 5/\(10^5\), 6/\(10^6\), 7/\(10^7\), 8/\(10^8\), 9/\(10^9\), 10/\(10^{10}\), 11/\(10^{11}\)} {
  \draw[black!50] (-0.08, \i*\yf) -- (0.08, \i*\yf);
  \node[font=\scriptsize, text=black, left] at (-0.12, \i*\yf) {\lab};
}
\node[font=\footnotesize, text=black, rotate=90] at (-1.4, \ymax/2) {\(T(n)\) (steps)};
\draw[lv0, line width=1.6pt] plot[smooth] coordinates {
  (1*\xf, 0) (2*\xf, 0.301*\yf) (4*\xf, 0.602*\yf) (7*\xf, 0.845*\yf)
  (10*\xf, 1*\yf) (15*\xf, 1.176*\yf) (20*\xf, 1.301*\yf)
  (25*\xf, 1.398*\yf) (30*\xf, 1.477*\yf) (35*\xf, 1.544*\yf)
};
\draw[lv2, line width=1.6pt] plot[smooth] coordinates {
  (1*\xf, 0) (2*\xf, 0.714*\yf) (4*\xf, 1.427*\yf) (7*\xf, 2.003*\yf)
  (10*\xf, 2.370*\yf) (15*\xf, 2.787*\yf) (20*\xf, 3.083*\yf)
  (25*\xf, 3.314*\yf) (30*\xf, 3.501*\yf) (35*\xf, 3.660*\yf)
};
\draw[lv3, line width=1.6pt] plot[smooth] coordinates {
  (1*\xf, 0) (2*\xf, 1.806*\yf) (4*\xf, 3.612*\yf) (7*\xf, 5.07*\yf)
  (10*\xf, 6*\yf) (15*\xf, 7.057*\yf) (20*\xf, 7.806*\yf)
  (25*\xf, 8.388*\yf) (30*\xf, 8.863*\yf) (35*\xf, 9.266*\yf)
};
\draw[lv5, line width=1.6pt] plot[smooth] coordinates {
  (1*\xf, 0.301*\yf) (2*\xf, 1.204*\yf) (3*\xf, 2.709*\yf)
  (4*\xf, 4.816*\yf) (5*\xf, 7.526*\yf) (5.5*\xf, 9.108*\yf)
  (6*\xf, 10.836*\yf)
};
\node[fill=white, draw=black!20, rounded corners=2pt, inner sep=5pt,
      anchor=north west, font=\footnotesize] at (0.3, \ymax-0.2) {
  \begin{tabular}{ll}
    \textcolor{lv0}{\rule[0.3ex]{0.8cm}{1.6pt}} & \(k=3\) Regular: \(\Theta(n)\) \\[1.5pt]
    \textcolor{lv2}{\rule[0.3ex]{0.8cm}{1.6pt}} & \(k=2\) CFG: \(O(n^\omega) \approx O(n^{2.37})\) \\[1.5pt]
    \textcolor{lv3}{\rule[0.3ex]{0.8cm}{1.6pt}} & \(k=2^+\) TAG/MCFG: \(O(n^6)\) \\[1.5pt]
    \textcolor{lv5}{\rule[0.3ex]{0.8cm}{1.6pt}} & \(k=1\) CSG: \(O(2^{n^2})\) \\[1.5pt]
    {\color{lv6}\rule[0.3ex]{0.4cm}{1pt}\hspace{2pt}\rule[0.3ex]{0.4cm}{1pt}} & \(k=0\) RE: undecidable \\
  \end{tabular}
};
\end{tikzpicture}
}
\caption{Semi-logarithmic scale of the best known algorithmic bound for the membership decision, class by class of the Chomsky hierarchy. The polynomial bounds for \(k=3\) (\(\Theta(n)\), green), \(k=2\) (\(O(n^{2.37})\), yellow) and \(k=2^+\) (\(O(n^6)\), orange) appear as straight lines of increasing slope; the slope is 1, 2.37 and 6 respectively on the semi-log scale. The exponential bound \(O(2^{n^2})\) for \(k=1\) CSG (magenta) diverges from \(n \approx 6\) beyond the represented scale. For \(k=0\) RE, undecidability is not representable on the \(n\) axis (complexity is not a function of input length); see the \(k=0\) paragraph for the detailed discussion.}
\label{fig:d1-decision}
\end{figure}

\textbf{Construction of a derivation tree.} Once the membership decision has been answered in the affirmative, one may wish to produce a structural witness --- a derivation tree \(t\) whose yield gives exactly \(w\). The fundamental result that structures this subsection is that constructing such a witness \textbf{costs only an additional constant factor over the membership decision}, throughout the hierarchy. The reason is uniform: the standard decision algorithms maintain, during their execution, an auxiliary structure (generally called a \emph{back-pointer table} or \emph{chart}) that records, for each recognized substring, the production rule and the subtrees that enabled that recognition. Once the table is complete, a backward traversal from the root --- starting from the cell that attests to the recognition of the whole string by the axiom --- reconstructs the tree in time linear in the size of that tree (at most \(O(n)\) internal nodes for a grammar in Chomsky normal form).

The concrete realization of this backward traversal varies by algorithm. For a deterministic finite automaton (\(k=3\)), no table is even needed: the sequence of states traversed during the DFA simulation already constitutes the tree (which here is a linear chain of rule applications). For CFGs (\(k=2\)), CYK \citep{younger1967recognition} fills a triangular table of size \(O(n^2)\) each of whose cells keeps pointers to its two sub-cells; the backward traversal reconstructs the tree in \(O(n)\) after the \(O(n^3)\) of the decision. Earley's algorithm \citep{earley1970efficient} maintains an item graph that, after the recognition phase, can be traversed to extract a derivation. Valiant's sub-cubic algorithm \citep{valiant1975general} relies on the transitive closure of a Boolean matrix whose coefficients record the rules used, from which a tree can be extracted. For TAG/MCFG (\(k=2^+\)), the tabular methods of \citet[§10.5]{joshischabes1997tree} produce a \emph{parse forest} in \(O(|G| \cdot n^6)\) from which extracting one tree takes an additional \(O(n)\). For CSGs (\(k=1\)), the situation is different: the decision is made by enumerating configurations reachable in space \(n^2\), and extracting a witness of acceptance is done by reconstituting the path traversed in that configuration space --- still in \(O(2^{n^2})\) time and \(O(n^2)\) space, like the decision. For recursively enumerable languages (\(k=0\)), finally, the tree-construction problem inherits the undecidability of the decision.

An important subtlety arises for \textbf{ambiguous} CFGs (and beyond). When a string \(w\) admits several valid derivation trees --- for example the multiple prepositional attachment in \emph{``I saw the man with the telescope''} --- constructing \textbf{one} tree requires a \emph{choice} among these alternatives. The cost of producing this single witness remains asymptotically the same as the membership decision, but the algorithm must resolve the non-determinism at some point, generally by an arbitrary deterministic strategy (take the first derivation found, or the leftmost, or the one that minimizes a score). For deterministic context-free languages (DCFLs), which are by definition unambiguous, this choice does not exist and the witness is unique --- which makes these subclasses particularly attractive for programming languages, where one wants a unique interpretation of each program.

\textbf{Enumeration of all trees.} The problem becomes qualitatively different as soon as one seeks to list \textbf{all} the valid derivation trees of a string \(w\), rather than a single witness. The cost can then explode combinatorially because it is output-bounded: at a minimum, each tree must be produced. The bounds vary considerably by class.

For a \textbf{regular} language (\(k=3\)), enumerating all trees is trivial: a deterministic finite automaton produces only \textbf{one} accepting path per recognized string (finite automata have no structural ambiguity in the sense of derivation trees). Enumeration therefore coincides with the construction of a single tree, in \(O(n)\).

For a \textbf{context-free} grammar (\(k=2\)), the number of valid derivation trees for an ambiguous string of length \(n\) grows asymptotically as the \(n\)-th \textbf{Catalan number},
\[C_n \;=\; \frac{1}{n+1}\binom{2n}{n} \;\sim\; \frac{4^n}{n^{3/2}\sqrt{\pi}}.\]
The Catalan numbers form a fundamental combinatorial sequence whose first terms are \(C_0 = 1, C_1 = 1, C_2 = 2, C_3 = 5, C_4 = 14, C_5 = 42, C_6 = 132, \ldots\) This sequence counts many equivalent combinatorial structures --- notably the number of full binary trees with \(n+1\) leaves, the number of ways to parenthesize a non-associative product of \(n+1\) terms (for example, \(C_3 = 5\) corresponds to the 5 distinct parenthesizations of \(a \cdot b \cdot c \cdot d\): \(((ab)c)d\), \((a(bc))d\), \((ab)(cd)\), \(a((bc)d)\), \(a(b(cd))\)), or the number of monotone lattice paths avoiding the diagonal in an \(n \times n\) grid. This sequence arises naturally in CFG parsing because each binary derivation tree in Chomsky normal form corresponds precisely to one way of grouping the input terminals into successive sub-constituents --- in other words, to a parenthesization of the input string. This bound manifests concretely in prepositional-attachment ambiguities: a sentence containing \(k\) successive prepositional phrases admits \(C_k\) possible attachment trees (for example \(C_3 = 5\), \(C_4 = 14\), \(C_{10} = 16\,796\)). Naive enumeration is therefore \(O(4^n)\). \citet{billot1989shared} showed that the set of these trees can be compactly represented by a \emph{shared parse forest} --- a directed acyclic graph that factorizes the subtrees common to several derivations --- in only \(O(n^3)\) space, the same complexity as the CYK table itself. This compact representation allows many queries (existence of at least one tree, counting, probabilistic sampling) to be answered in polynomial time, but the \textbf{explicit extraction} of individual trees from the forest remains of complexity \(O(4^n)\) in the worst case, simply because the number of trees to list can be effectively exponential.

For \textbf{mildly context-sensitive} grammars (\(k=2^+\)), the situation is analogous: the number of TAG or MCFG trees can reach \(O(4^n)\) or worse depending on the formalism and the nature of the adjunction or composition rules allowed. Shared-forest techniques extend to these formalisms (a generalization of Billot--Lang to \emph{shared parse forests} for TAG), permitting a polynomial representation of the forest (\(O(n^6)\) for TAG, in parallel with the decision bound), but explicit extraction remains exponential in the worst case.

For \textbf{context-sensitive} grammars (\(k=1\)), enumeration inherits the complexity of the reachable-configuration space. Each configuration of a nondeterministic linear-bounded automaton (cf.~Kuroda 1964) can correspond to a distinct derivation, and the total number of reachable configurations is bounded by \(2^{O(n^2)}\) (Savitch's theorem). Enumerating all trees is therefore \(O(2^{n^2})\) --- the same bound as the membership decision, because for CSGs the difficulty lies in exploring the configuration space, and enumerating all accepting paths does not change the asymptotic order of that exploration.

For \textbf{recursively enumerable} languages (\(k=0\)), finally, enumeration inherits the undecidability of the decision: if one cannot even determine whether a tree exists, one can a fortiori not list them all. More subtly, even for strings belonging to the language, the number of trees can be infinite (a non-monotonic RE grammar can admit arbitrarily long derivations for the same string by reintroducing non-terminals).

\begin{table}[H]
\centering
\small
\caption{Bounds for enumerating all trees, by class.}
\label{tab:d1-enum}
\begin{tabular}{@{}l p{0.22\columnwidth} p{0.30\columnwidth} p{0.26\columnwidth}@{}}
\toprule
Class & Explicit enumeration & Compact representation & Characterization \\
\midrule
\(k=3\) Regular & \(O(n)\) & \(O(n)\) (single tree) & tight, no structural ambiguity \\
\(k=2\) CFG & \(O(4^n)\) (Catalan) & \(O(n^3)\) shared forest (Billot \& Lang 1989) & tight (asymptotic Catalan) \\
\(k=2^+\) TAG/MCFG & \(O(4^n)\) or worse & \(O(n^6)\) shared forest (Billot--Lang generalization) & conditional BMM + combinatorial explosion \\
\(k=1\) CSG & \(O(2^{n^2})\) & n/a (same bound) & identical to the decision (PSPACE) \\
\(k=0\) RE & undecidable / infinite & n/a & undecidable + potentially infinite number of trees \\
\bottomrule
\end{tabular}
\end{table}

\begin{figure}[H]
\centering
\resizebox{\columnwidth}{!}{%
\begin{tikzpicture}[scale=1.3, transform shape]
\def\xmax{13}
\def\ymax{8.5}
\def\xf{0.3714}
\def\yf{0.7727}
\foreach \i in {0,1,...,11} {
  \draw[black!12] (0,\i*\yf) -- (\xmax,\i*\yf);
}
\foreach \i in {5,10,15,20,25,30,35} {
  \draw[black!12] (\i*\xf,0) -- (\i*\xf,\ymax);
}
\draw[black!60, line width=0.6pt, -{latex}] (0,0) -- (\xmax+0.4,0);
\draw[black!60, line width=0.6pt, -{latex}] (0,0) -- (0,\ymax+0.4);
\foreach \v in {5,10,15,20,25,30,35} {
  \draw[black!50] (\v*\xf, -0.08) -- (\v*\xf, 0.08);
  \node[font=\scriptsize, text=black, below] at (\v*\xf, -0.1) {\v};
}
\node[font=\footnotesize, text=black] at (\xmax/2, -0.7) {\(n\) (input length)};
\foreach \i/\lab in {0/\(1\), 1/\(10\), 2/\(10^2\), 3/\(10^3\), 4/\(10^4\), 5/\(10^5\), 6/\(10^6\), 7/\(10^7\), 8/\(10^8\), 9/\(10^9\), 10/\(10^{10}\), 11/\(10^{11}\)} {
  \draw[black!50] (-0.08, \i*\yf) -- (0.08, \i*\yf);
  \node[font=\scriptsize, text=black, left] at (-0.12, \i*\yf) {\lab};
}
\node[font=\footnotesize, text=black, rotate=90] at (-1.4, \ymax/2) {\(T(n)\) (steps)};
\draw[lv0, line width=1.6pt] plot[smooth] coordinates {
  (1*\xf, 0) (2*\xf, 0.301*\yf) (4*\xf, 0.602*\yf) (7*\xf, 0.845*\yf)
  (10*\xf, 1*\yf) (15*\xf, 1.176*\yf) (20*\xf, 1.301*\yf)
  (25*\xf, 1.398*\yf) (30*\xf, 1.477*\yf) (35*\xf, 1.544*\yf)
};
\draw[lv4b, line width=1.6pt, densely dashed] plot[smooth] coordinates {
  (1*\xf, 0.602*\yf) (2*\xf, 1.204*\yf) (3*\xf, 1.806*\yf)
  (5*\xf, 3.010*\yf) (7*\xf, 4.214*\yf) (10*\xf, 6.021*\yf)
  (13*\xf, 7.826*\yf) (15*\xf, 9.031*\yf) (18*\xf, 10.836*\yf)
};
\draw[lv5, line width=1.6pt] plot[smooth] coordinates {
  (1*\xf, 0.301*\yf) (2*\xf, 1.204*\yf) (3*\xf, 2.709*\yf)
  (4*\xf, 4.816*\yf) (5*\xf, 7.526*\yf) (5.5*\xf, 9.108*\yf)
  (6*\xf, 10.836*\yf)
};
\node[fill=white, draw=black!20, rounded corners=2pt, inner sep=5pt,
      anchor=north west, font=\footnotesize] at (0.3, \ymax-0.2) {
  \begin{tabular}{ll}
    \textcolor{lv0}{\rule[0.3ex]{0.8cm}{1.6pt}} & \(k=3\) Regular: \(O(n)\) \\[1.5pt]
    {\color{lv4b}\rule[0.3ex]{0.18cm}{1.6pt}\hspace{1pt}\rule[0.3ex]{0.18cm}{1.6pt}\hspace{1pt}\rule[0.3ex]{0.18cm}{1.6pt}} & \(k\!=\!2, 2^+\) CFG/TAG: \(O(4^n)\) (Catalan) \\[1.5pt]
    \textcolor{lv5}{\rule[0.3ex]{0.8cm}{1.6pt}} & \(k=1\) CSG: \(O(2^{n^2})\) \\[1.5pt]
    {\color{lv6}\rule[0.3ex]{0.4cm}{1pt}\hspace{2pt}\rule[0.3ex]{0.4cm}{1pt}} & \(k=0\) RE: undecidable / infinite \\
  \end{tabular}
};
\end{tikzpicture}
}
\caption{Semi-logarithmic scale of the complexity of the \textbf{explicit} enumeration of all derivation trees, class by class. To be compared with Figure~\ref{fig:d1-decision} (membership decision): for \(k=3\) the curve is identical (\(O(n)\), one tree per string), but from \(k=2\) onward the bound \(O(4^n)\) (Catalan numbers, a line of slope \(\log_{10}(4) \approx 0.602\) on the semi-log scale) \textbf{diverges far faster} than the \(O(n^{2.37})\) bound of the decision --- a concrete illustration of the ``second dimension of difficulty'' introduced by the combinatorial explosion of the set of trees to produce. For \(k=1\) and \(k=0\), the bounds coincide with those of the membership decision (enumeration inherits, respectively, the exploration of the configuration space and the undecidability). The \(O(4^n)\) curve overtakes the CFG-decision curve (\(O(n^{2.37})\)) as early as \(n \approx 5\), and reaches \(10^{11}\) steps at \(n = 18\) already.}
\label{fig:d1-enum}
\end{figure}

\subsubsection*{Generation side}

\textbf{Free generation.} The first of the three sub-problems on the generation side consists in applying the grammar's production rules with no particular target, stopping at an arbitrary step. The cost is then measured in \emph{derivation steps} (each step rewrites one non-terminal according to a rule), not in total elementary operations --- this is the natural measure for assessing the intrinsic cost of the rewriting operation, independent of the cost of manipulating auxiliary structures.

For a \textbf{regular} language (\(k=3\)), free generation is performed by walking through the DFA (deterministic or not): each step corresponds to reading a transition, in \(O(1)\); the total cost of producing a form of length \(n\) is \(\Theta(n)\) \citep[ch.~2--4]{hopcroft2006introduction}. Termination is guaranteed as soon as a final state is reachable from every state of the DFA.

For a \textbf{context-free} grammar (\(k=2\)), the leftmost or rightmost derivation proceeds by rewriting, at each step, the leftmost (or rightmost) non-terminal of the current sentential form. The cost per step is \(O(1)\) if the non-terminal frontier is maintained explicitly (by a queue or stack), \(O(n)\) with a naive scan of the sentential form. The total cost is therefore \(O(n)\) to \(O(n^2)\) for \(n\) steps \citep{earley1970efficient}; \citep[Vol.~I]{aho1986compilers}. A subtlety: for grammars containing unit rules (\(A \to B\)), the derivation can enter a cycle without producing a terminal if the rewriting order is poorly controlled; standard implementations forbid or eliminate these unit rules.

For \textbf{mildly context-sensitive} grammars (\(k=2^+\)), the analogue of derivation is the \emph{adjunction} operation (TAG) or \emph{composition} operation (MCFG). \citet[§6.1]{joshi1985formalisme} describes TAG adjunction as the insertion of an auxiliary tree \(\beta\) at a non-terminal node of a current tree \(\gamma\). The cost per step is \(O(|G|)\) for matching the local adjunction constraints. For \(n\) steps, the total cost is of the order of \(O(n \cdot |G|)\), i.e.\ polynomial for a fixed grammar.

For \textbf{context-sensitive} grammars (\(k=1\)), each derivation step applies a rule of the form \(\alpha A \beta \to \alpha \gamma \beta\), which requires a \emph{context match} in the current sentential form. Naive matching is \(O(n^2)\) per step (scanning the sentential form to identify all positions where the context \(\alpha A \beta\) appears). For \(n\) steps, the total cost reaches \(O(n^3)\). Termination is guaranteed for monotonic grammars (each rule increases or preserves the length of the sentential form), as \citet{kuroda1964classes} observes in his analysis of linear-bounded automata.

For \textbf{recursively enumerable} languages (\(k=0\)), non-monotonic rules (length-reducing rules, such as \(\alpha A \beta \to \gamma\) with \(|\gamma| < |\alpha A \beta|\)) can rewrite the sentential form indefinitely without converging to a terminal string. The derivation may therefore fail to terminate, and the cost is unbounded --- as for the membership decision, there is no computable function bounding the termination time in the cases where the derivation eventually produces a complete string.

\textbf{Terminating example-generation.} The second sub-problem requires the derivation to produce a complete string of \(L(G)\), that is, a sentential form consisting exclusively of terminal symbols. The relevant measure then becomes the \emph{total elementary operations}, rule applications included, as a function of the length \(n\) of the produced string.

For a \textbf{regular} language (\(k=3\)), a string of length \(n\) is obtained by executing \(n\) DFA transitions up to a final state. Total cost \(\Theta(n)\): example-generation coincides with the membership decision, with no asymmetry whatsoever.

For a \textbf{context-free} grammar (\(k=2\)), the derivation producing a string of length \(n\) requires at most \(n\) steps in Chomsky normal form (each step produces exactly two symbols, of which at least one is a terminal at the leaf level). With a non-terminal frontier maintained explicitly, the cost per step is \(O(1)\) and the total cost \(O(n)\). With a naive scan, the total cost rises to \(O(n^2)\) \citep{younger1967recognition}; \citep{hopcroft2006introduction}.

For \textbf{mildly context-sensitive} grammars (\(k=2^+\)), generating a TAG or MCFG string of length \(n\) by top-down derivation over the elementary trees costs \(O(n^2)\) to \(O(n^3)\) depending on the formalism and the strategy for choosing auxiliary trees \citep{joshischabes1997tree}; \citep{vijayshankerweir1994equivalence}. No intrinsic tight characterization is published for this sub-problem --- the literature on TAG/MCFG generation complexity is appreciably less developed than that on the corresponding recognition.

For \textbf{context-sensitive} grammars (\(k=1\)), the derivation produces a string of length \(n\) in at most \(n\) steps (by monotonicity, \citealt{kuroda1964classes}), each step costing \(O(n^2)\) for naive context matching. Total cost \(O(n^3)\). Example-generation therefore remains \emph{polynomial} for CSGs, in sharp contrast with the membership decision, which is PSPACE-complete --- this is the first strong manifestation of the asymmetry in favor of generation at this level of the hierarchy.

For \textbf{recursively enumerable} languages (\(k=0\)), terminating example-generation is \emph{semi-decidable}: if a string \(w \in L(G)\) exists, a finite derivation exists; but no computable bound exists a priori on its length. The algorithm enumerating the possible derivations halts in finite time for \(w \in L(G)\), but this time is not bounded by any recursive function of \(n\).

\begin{table}[H]
\centering
\small
\caption{Bounds for terminating example-generation, by class.}
\label{tab:d1-genex}
\begin{tabular}{@{}l l p{0.42\columnwidth}@{}}
\toprule
Class & Total cost & Characterization \\
\midrule
\(k=3\) Regular & \(\Theta(n)\) & tight, identical to the membership decision \\
\(k=2\) CFG & \(O(n)\) to \(O(n^2)\) & in \(\mathsf{P}\), not tight \\
\(k=2^+\) TAG/MCFG & \(O(n^2)\) to \(O(n^3)\) & in \(\mathsf{P}\), not tight \\
\(k=1\) CSG & \(O(n^3)\) & in \(\mathsf{P}\), sharp contrast with PSPACE-completeness of the decision \\
\(k=0\) RE & unbounded (semi-decidable) & undecidable in the strict sense \\
\bottomrule
\end{tabular}
\end{table}

\begin{figure}[H]
\centering
\resizebox{\columnwidth}{!}{%
\begin{tikzpicture}[scale=1.3, transform shape]
\def\xmax{13}
\def\ymax{8.5}
\def\xf{0.3714}
\def\yf{0.7727}
\foreach \i in {0,1,...,11} {
  \draw[black!12] (0,\i*\yf) -- (\xmax,\i*\yf);
}
\foreach \i in {5,10,15,20,25,30,35} {
  \draw[black!12] (\i*\xf,0) -- (\i*\xf,\ymax);
}
\draw[black!60, line width=0.6pt, -{latex}] (0,0) -- (\xmax+0.4,0);
\draw[black!60, line width=0.6pt, -{latex}] (0,0) -- (0,\ymax+0.4);
\foreach \v in {5,10,15,20,25,30,35} {
  \draw[black!50] (\v*\xf, -0.08) -- (\v*\xf, 0.08);
  \node[font=\scriptsize, text=black, below] at (\v*\xf, -0.1) {\v};
}
\node[font=\footnotesize, text=black] at (\xmax/2, -0.7) {\(n\) (length of the produced string)};
\foreach \i/\lab in {0/\(1\), 1/\(10\), 2/\(10^2\), 3/\(10^3\), 4/\(10^4\), 5/\(10^5\), 6/\(10^6\), 7/\(10^7\), 8/\(10^8\), 9/\(10^9\), 10/\(10^{10}\), 11/\(10^{11}\)} {
  \draw[black!50] (-0.08, \i*\yf) -- (0.08, \i*\yf);
  \node[font=\scriptsize, text=black, left] at (-0.12, \i*\yf) {\lab};
}
\node[font=\footnotesize, text=black, rotate=90] at (-1.4, \ymax/2) {\(T(n)\) (steps)};
\draw[lv0, line width=1.6pt] plot[smooth] coordinates {
  (1*\xf, 0) (2*\xf, 0.301*\yf) (4*\xf, 0.602*\yf) (7*\xf, 0.845*\yf)
  (10*\xf, 1*\yf) (15*\xf, 1.176*\yf) (20*\xf, 1.301*\yf)
  (25*\xf, 1.398*\yf) (30*\xf, 1.477*\yf) (35*\xf, 1.544*\yf)
};
\draw[lv1, line width=1.6pt] plot[smooth] coordinates {
  (1*\xf, 0) (2*\xf, 0.602*\yf) (4*\xf, 1.204*\yf) (7*\xf, 1.690*\yf)
  (10*\xf, 2*\yf) (15*\xf, 2.352*\yf) (20*\xf, 2.602*\yf)
  (25*\xf, 2.796*\yf) (30*\xf, 2.954*\yf) (35*\xf, 3.088*\yf)
};
\draw[lv2, line width=1.6pt] plot[smooth] coordinates {
  (1*\xf, 0) (2*\xf, 0.903*\yf) (4*\xf, 1.806*\yf) (7*\xf, 2.535*\yf)
  (10*\xf, 3*\yf) (15*\xf, 3.528*\yf) (20*\xf, 3.903*\yf)
  (25*\xf, 4.194*\yf) (30*\xf, 4.431*\yf) (35*\xf, 4.633*\yf)
};
\node[fill=white, draw=black!20, rounded corners=2pt, inner sep=5pt,
      anchor=north west, font=\footnotesize] at (0.3, \ymax-0.2) {
  \begin{tabular}{ll}
    \textcolor{lv0}{\rule[0.3ex]{0.8cm}{1.6pt}} & \(k=3\) Regular: \(\Theta(n)\) \\[1.5pt]
    \textcolor{lv1}{\rule[0.3ex]{0.8cm}{1.6pt}} & \(k=2\) CFG: \(O(n^2)\) \\[1.5pt]
    \textcolor{lv2}{\rule[0.3ex]{0.8cm}{1.6pt}} & \(k=2^+\) TAG/MCFG \& \(k=1\) CSG: \(O(n^3)\) \\[1.5pt]
    {\color{lv6}\rule[0.3ex]{0.4cm}{1pt}\hspace{2pt}\rule[0.3ex]{0.4cm}{1pt}} & \(k=0\) RE: unbounded (semi-dec.) \\
  \end{tabular}
};
\end{tikzpicture}
}
\caption{Semi-logarithmic scale of the complexity of terminating example-generation, class by class. The bounds remain polynomial for all monotonic classes (\(k=3\) to \(k=1\)), with moderate growth --- slope 1 for regular, slope 2 for CFGs, slope 3 for TAG/MCFG and CSG. To be compared with Figure~\ref{fig:d1-decision} (membership decision): generation is strictly faster than the decision for \(k=2^+\) and \(k=1\), and the gap becomes qualitative for \(k=1\) (polynomial vs PSPACE-complete). For \(k=0\), the derivation may fail to terminate; this case is not representable on the \(n\) axis (see legend).}
\label{fig:d1-genex}
\end{figure}

\textbf{Generation under semantic constraint.} The third sub-problem is qualitatively different from the two preceding ones. When the produced string must, in addition to belonging to \(L(G)\), satisfy an \emph{external constraint} \(C\) (typically a semantic representation, a pragmatic requirement, or a formal target), the complexity profile changes radically and can, in certain cases, exceed that of the membership decision --- this is what constitutes the ``sign reversal'' of the asymmetry, discussed at the end of this subsection.

For a \textbf{regular} language (\(k=3\)), the semantic constraint is typically encoded as a second finite automaton \(A_C\) that characterizes the strings satisfying \(C\). Generation under constraint then amounts to producing a string in the intersection language \(L(G) \cap L(A_C)\), computable by an automaton product in \(O(|Q_G| \cdot |Q_C|)\) states. A string of the intersection language is then generated in an additional \(\Theta(n)\). Total cost \(\Theta(n + |Q_G| \cdot |Q_C|)\), polynomial. \citet{mohri1997finitestate} establishes, in the more general framework of sequential finite-state transducers, that the output of a deterministic transducer:

\begin{quote}
\emph{``depends, in general linearly, only on the input size and can therefore be considered optimal from this point of view.''} \citep[p.~270]{mohri1997finitestate}
\end{quote}

The regular case therefore remains polynomial even under semantic constraint --- it is the only class where the three generation sub-problems coincide asymptotically.

For a \textbf{pure context-free} grammar (\(k=2\), with no \textbf{feature structures} --- attribute-value annotations attached to constituents to express agreement constraints (subject-verb, gender-number), subcategorization constraints (a transitive verb requires a complement), or long-distance dependencies --- nor other non-local constraints), generation under semantic constraint remains polynomial if the semantics is strictly compositional. But as soon as the grammar is enriched with feature structures or unification --- the standard case of modern linguistic grammars (LFG, HPSG, FTAG) --- the complexity explodes. \citet[§4]{kay1996chart} establishes that generation from a ``flat'' semantic representation is intrinsically exponential in the worst case:

\begin{quote}
\emph{``The process is exponential in the worst case because, if a sentence contains a word with \(k\) modifiers, then a version will be generated with each of the \(2^k\) subsets of those modifiers, all but one of them being rejected when it is finally discovered that their semantics does not subsume the entire input.''} \citep[p.~202]{kay1996chart}
\end{quote}

The structural cause is the absence of positional indexing: the logical indices of the semantic representation do not correspond to positions in the string, which removes the mechanism that makes chart-based parsing polynomial. The generator must consider all the compatible semantic sub-coverings, whose number is exponential. Beyond Kay's analysis, \citet{brew1992letting} \textbf{formally proves} the NP-completeness of a minimal scheme for CFG generation from a bag (multi-set) of lexical signs, by polynomial reduction from 3-Dimensional Matching:

\begin{quote}
\emph{``We now provide a polynomial-time reduction from an arbitrary instance of MENAGE A TROIS to an instance of Shake-and-Bake generation, which allows the same conclusion to be drawn for this problem.''} \citep[§2.1.3, p.~611]{brew1992letting}
\end{quote}

Brew's result is important because it proves \emph{intrinsic} NP-completeness (and not merely worst-case difficulty) on a minimal scheme that depends on no complex auxiliary structure. Generation under lexical-occurrence constraint therefore exceeds polynomial as early as \(k=2\).

For \textbf{mildly context-sensitive} grammars (\(k=2^+\)), no new tight bound is published for generation under semantic constraint. Kay's same exponential argument \citep{kay1996chart} extends to FTAG grammars enriched with feature structures, and one can anticipate that the NP-completeness established by Brew for CFGs transfers by direct reduction to TAGs, which are a strict extension of CFGs. \citet{russell1990asymmetry} confirm empirically, through two case studies (French clitics, \emph{empty semantic heads} in German-Dutch), that \emph{``all known methods of generation impose constraints on the grammars they assume''} \citep[abstract, p.~205]{russell1990asymmetry}, and conclude that generation with algorithms of this class often leads to non-termination on grammars that are correct for parsing \citep[p.~210]{russell1990asymmetry}.

For \textbf{context-sensitive} grammars (\(k=1\)), the intrinsic lower bound reaches at least NP-hard. \citet[Ch.~3, p.~93--95]{barton1987complexity} establish that the membership decision for \emph{Agreement Grammars} (a simplified model of agreement + ambiguity corresponding to CFGs enriched with limited feature structures) is NP-complete, by reduction from 3-SAT. This decision result transfers to generation under constraint by a trivial reduction: in order to produce a string satisfying the constraint, the generator must in particular be able to recognize which strings satisfy it, and hence solve the underlying decision problem. Generation under semantic constraint for CSGs is therefore at least NP-hard, and possibly PSPACE-hard by inheritance from the PSPACE-completeness of the CSG decision.

For \textbf{recursively enumerable} languages (\(k=0\)), generation under semantic constraint is \emph{undecidable} in general. Given an arbitrary semantic constraint \(C\), deciding whether there exists \(w \in L(G)\) satisfying \(C\) amounts to the halting problem \citep{turing1936computable}.

\begin{table}[H]
\centering
\small
\caption{Bounds for generation under semantic constraint, by class.}
\label{tab:d1-genconstraint}
\begin{tabular}{@{}p{0.42\columnwidth} p{0.30\columnwidth} l@{}}
\toprule
Class & Bound & Pivot source \\
\midrule
\(k=3\) Regular & \(\Theta(n + |Q_G| \cdot |Q_C|)\) & Mohri 1997 \\
\(k=2\) pure CFG (compositional semantics) & polynomial & folklore \\
\(k=2\) CFG + features/unification & exponential (\(2^k\) modifiers) & Kay 1996, §4 \\
\(k=2\) CFG + multi-set input & \textbf{NP-complete (proven)} & Brew 1992, §2.1.3 \\
\(k=2^+\) TAG/MCFG + features & exponential (Kay + Brew inheritance) & Kay 1996, Brew 1992 \\
\(k=1\) CSG + features & \textbf{NP-hard to PSPACE-hard} & Barton 1987, Ch.~3 \\
\(k=0\) RE & \textbf{undecidable} & Turing 1936 \\
\bottomrule
\end{tabular}
\end{table}

\begin{figure}[H]
\centering
\resizebox{\columnwidth}{!}{%
\begin{tikzpicture}[scale=1.3, transform shape]
\def\xmax{13}
\def\ymax{8.5}
\def\xf{0.3714}
\def\yf{0.7727}
\foreach \i in {0,1,...,11} {
  \draw[black!12] (0,\i*\yf) -- (\xmax,\i*\yf);
}
\foreach \i in {5,10,15,20,25,30,35} {
  \draw[black!12] (\i*\xf,0) -- (\i*\xf,\ymax);
}
\draw[black!60, line width=0.6pt, -{latex}] (0,0) -- (\xmax+0.4,0);
\draw[black!60, line width=0.6pt, -{latex}] (0,0) -- (0,\ymax+0.4);
\foreach \v in {5,10,15,20,25,30,35} {
  \draw[black!50] (\v*\xf, -0.08) -- (\v*\xf, 0.08);
  \node[font=\scriptsize, text=black, below] at (\v*\xf, -0.1) {\v};
}
\node[font=\footnotesize, text=black] at (\xmax/2, -0.7) {\(n\) (constraint size or target string)};
\foreach \i/\lab in {0/\(1\), 1/\(10\), 2/\(10^2\), 3/\(10^3\), 4/\(10^4\), 5/\(10^5\), 6/\(10^6\), 7/\(10^7\), 8/\(10^8\), 9/\(10^9\), 10/\(10^{10}\), 11/\(10^{11}\)} {
  \draw[black!50] (-0.08, \i*\yf) -- (0.08, \i*\yf);
  \node[font=\scriptsize, text=black, left] at (-0.12, \i*\yf) {\lab};
}
\node[font=\footnotesize, text=black, rotate=90] at (-1.4, \ymax/2) {\(T(n)\) (steps)};
\draw[lv0, line width=1.6pt] plot[smooth] coordinates {
  (1*\xf, 0) (2*\xf, 0.301*\yf) (4*\xf, 0.602*\yf) (7*\xf, 0.845*\yf)
  (10*\xf, 1*\yf) (15*\xf, 1.176*\yf) (20*\xf, 1.301*\yf)
  (25*\xf, 1.398*\yf) (30*\xf, 1.477*\yf) (35*\xf, 1.544*\yf)
};
\draw[lv1, line width=1.6pt, densely dashed] plot[smooth] coordinates {
  (1*\xf, 0) (2*\xf, 0.602*\yf) (4*\xf, 1.204*\yf) (7*\xf, 1.690*\yf)
  (10*\xf, 2*\yf) (15*\xf, 2.352*\yf) (20*\xf, 2.602*\yf)
  (25*\xf, 2.796*\yf) (30*\xf, 2.954*\yf) (35*\xf, 3.088*\yf)
};
\draw[lv4, line width=1.6pt] plot[smooth] coordinates {
  (1*\xf, 0.301*\yf) (3*\xf, 0.903*\yf) (5*\xf, 1.505*\yf)
  (10*\xf, 3.010*\yf) (15*\xf, 4.515*\yf) (20*\xf, 6.021*\yf)
  (25*\xf, 7.526*\yf) (30*\xf, 9.031*\yf) (35*\xf, 10.536*\yf)
};
\draw[lv5, line width=1.6pt, densely dotted] plot[smooth] coordinates {
  (1*\xf, 0.301*\yf) (2*\xf, 1.204*\yf) (3*\xf, 2.709*\yf)
  (4*\xf, 4.816*\yf) (5*\xf, 7.526*\yf) (5.5*\xf, 9.108*\yf)
  (6*\xf, 10.836*\yf)
};
\node[fill=white, draw=black!20, rounded corners=2pt, inner sep=5pt,
      anchor=north west, font=\footnotesize] at (0.3, \ymax-0.2) {
  \begin{tabular}{ll}
    \textcolor{lv0}{\rule[0.3ex]{0.8cm}{1.6pt}} & \(k=3\) Regular: \(\Theta(n + |Q_G||Q_C|)\) \\[1.5pt]
    {\color{lv1}\rule[0.3ex]{0.18cm}{1.6pt}\hspace{1pt}\rule[0.3ex]{0.18cm}{1.6pt}\hspace{1pt}\rule[0.3ex]{0.18cm}{1.6pt}} & \(k=2\) pure CFG: polynomial \\[1.5pt]
    \textcolor{lv4}{\rule[0.3ex]{0.8cm}{1.6pt}} & \(k\!=\!2\) features/multi-set, \(k\!=\!2^+\): \(O(2^n)\) NP \\[1.5pt]
    {\color{lv5}\rule[0.3ex]{0.1cm}{1.6pt}\hspace{1.5pt}\rule[0.3ex]{0.1cm}{1.6pt}\hspace{1.5pt}\rule[0.3ex]{0.1cm}{1.6pt}\hspace{1.5pt}\rule[0.3ex]{0.1cm}{1.6pt}} & \(k=1\) CSG features: \(O(2^{n^2})\) NP-hard \\[1.5pt]
    {\color{lv6}\rule[0.3ex]{0.4cm}{1pt}\hspace{2pt}\rule[0.3ex]{0.4cm}{1pt}} & \(k=0\) RE: undecidable \\
  \end{tabular}
};
\end{tikzpicture}
}
\caption{Semi-logarithmic scale of the complexity of generation under semantic constraint, class by class and sub-case by sub-case. Polynomial bounds persist for regular and pure CFGs, but the introduction of feature structures or non-local constraints tips the complexity into exponential as early as \(k=2\) (\(O(2^n)\) NP-complete, proven by Brew 1992) or into NP-hard to PSPACE-hard at \(k=1\). To be compared with Figure~\ref{fig:d1-decision} (membership decision): here generation \textbf{exceeds} the decision in complexity from \(k=2\) onward with features, whereas in Figure~\ref{fig:d1-decision} the decision remained polynomial up to \(k=2^+\). This is the visual manifestation of the \emph{sign reversal} of the asymmetry under semantic constraint.}
\label{fig:d1-genconstraint}
\end{figure}

\textbf{The sign reversal of the asymmetry.} The most striking result of this generation-side analysis is the \textbf{sign reversal} of the asymmetry for generation under semantic constraint. On the recognition side, complexity grows with position in the hierarchy (\(\Theta(n)\) \(\to\) \(O(n^\omega)\) \(\to\) \(O(n^6)\) \(\to\) PSPACE-complete \(\to\) undecidable). On the generation side, free generation and terminating example-generation remain polynomial up to and including \(k=1\) (at most \(O(n^3)\)). But for generation under semantic constraint, the asymmetry changes sign as early as \(k=2\) with features/unification: generation becomes exponential \citep{kay1996chart} while recognition remains polynomial; the reversal is even \emph{formally proven} on a minimal scheme (CFG + multi-set input) by \citet{brew1992letting} via reduction from 3-Dimensional Matching, which proves \textbf{intrinsic} NP-completeness while the membership decision remains in \(\mathsf{P}\). This finding invalidates the common rhetorical characterization ``generation is easy, parsing is hard'': generation under semantic constraint can be strictly harder than recognition, and the asymmetry changes nature according to the sub-problem considered, not only according to the class. This discussion reappears in §6 in connection with constrained generation by large language models, which embody this reversal in modern architectures (LLMs constrained on format, factuality, safety).

\subsubsection*{Synthesis: differential coupling}

The bounds established in the two preceding subsections can now be set against each other to bring out what constitutes the core of D1: a \textbf{differential coupling} between the two sides and their sub-problems. For recognition, it is the \emph{grammatical class} that dominates the complexity --- the membership decision and the construction of a tree follow the same curve along the hierarchy (\(\Theta(n)\) for regular, \(O(n^\omega)\) for CFGs, \(O(n^6)\) for TAG/MCFG, PSPACE-complete for CSGs, undecidable for RE), and only the enumeration of all trees introduces a second dimension of difficulty tied to the combinatorics of the number of trees (\(O(4^n)\) for ambiguous CFGs, via the Catalan numbers). For generation, conversely, it is the \emph{sub-problem} that dominates --- free generation and terminating example-generation remain polynomial across all monotonic classes, but generation under semantic constraint tips into NP-complete as early as \(k=2\) and stays there everywhere above. Table~\ref{tab:d1-compare} below first compares the two most contrasted focal cases (membership decision on the recognition side, terminating example-generation on the generation side), then the two heatmaps of Figure~\ref{fig:heatmaps} visualize this coupling across the six complete sub-problems.

\begin{table}[H]
\centering
\small
\caption{Comparison between the membership decision (recognition side) and terminating example-generation (generation side), by class.}
\label{tab:d1-compare}
\begin{tabular}{@{}l l l p{0.34\columnwidth}@{}}
\toprule
Class & Membership decision & Example-generation & Gap between the two \\
\midrule
\(k=3\) & \(\Theta(n)\) & \(O(n)\) & none \\
\(k=2\) & \(O(n^3)\) to \(O(n^{2.37})\) & \(O(n)\) to \(O(n^2)\) & polynomial, from \(n\) to \(n^{0.37}\) \\
\(k=2^+\) & \(O(n^6)\) & \(O(n^2)\) to \(O(n^3)\) & polynomial, from \(n^3\) to \(n^4\) \\
\(k=1\) & \(O(2^{n^2})\) (PSPACE-complete) & \(O(n^3)\) & class separation (\(\mathsf{P}\) vs \(\mathsf{PSPACE}\)) \\
\(k=0\) & undecidable & unbounded & decidability separation \\
\bottomrule
\end{tabular}
\end{table}

\begin{figure}[H]
\centering
\resizebox{0.92\columnwidth}{!}{%
\begin{tikzpicture}[scale=1.5, transform shape,
  hcell/.style={minimum width=2.3cm, minimum height=1cm, anchor=center,
    font=\footnotesize\bfseries, text=black, rounded corners=3pt, line width=0.4pt}]
\node[font=\large\bfseries, text=black] at (5.5,4.6) {Generation complexity};
\node[font=\scriptsize, text=black!60] at (5.5,4.15) {(sub-problem \(\times\) grammar class)};
\foreach \x/\lab in {1.1/k\!=\!3, 3.3/k\!=\!2, 5.5/k\!=\!2^+, 7.7/k\!=\!1, 9.9/k\!=\!0} {
  \node[font=\footnotesize\bfseries, text=black] at (\x,3.55) {\(\lab\)};
}
\node[font=\footnotesize\bfseries, text=black, anchor=east, align=right] at (-0.15,2.7) {Free};
\node[font=\footnotesize\bfseries, text=black, anchor=east, align=right] at (-0.15,1.6) {Terminating \\ example};
\node[font=\footnotesize\bfseries, text=black, anchor=east, align=right] at (-0.15,0.5) {Under semantic \\ constraint};
\foreach \x in {1.1,3.3,5.5,7.7} {
  \node[hcell, fill=lv0!25, draw=lv0!60] at (\x,2.7) {\(O(n)\)};
}
\node[hcell, fill=lv6!30, draw=lv6!70, text=black] at (9.9,2.7) {unbnd.};
\node[hcell, fill=lv0!25, draw=lv0!60] at (1.1,1.6) {\(O(n)\)};
\node[hcell, fill=lv1!30, draw=lv1!60] at (3.3,1.6) {\(O(n^2)\)};
\node[hcell, fill=lv2!40, draw=lv2!70] at (5.5,1.6) {\(O(n^3)\)};
\node[hcell, fill=lv2!40, draw=lv2!70] at (7.7,1.6) {\(O(n^3)\)};
\node[hcell, fill=lv6!30, draw=lv6!70, text=black] at (9.9,1.6) {unbnd.};
\node[hcell, fill=lv1!30, draw=lv1!60] at (1.1,0.5) {\(O(n^c)\)};
\node[hcell, fill=lv4!40, draw=lv4!70] at (3.3,0.5) {NP, \(O(2^n)\)};
\node[hcell, fill=lv4!40, draw=lv4!70] at (5.5,0.5) {NP};
\node[hcell, fill=lv5!35, draw=lv5!70, text=black] at (7.7,0.5) {\(O(2^{n^2})\)};
\node[hcell, fill=lv6!30, draw=lv6!70, text=black] at (9.9,0.5) {undec.};
\foreach \y in {3.2, 2.15, 1.05, -0.05} {
  \draw[black!20, line width=0.3pt] (0,\y) -- (11,\y);
}
\end{tikzpicture}
}

\medskip

\resizebox{0.92\columnwidth}{!}{%
\begin{tikzpicture}[scale=1.5, transform shape,
  hcell/.style={minimum width=2.3cm, minimum height=1cm, anchor=center,
    font=\footnotesize\bfseries, text=black, rounded corners=3pt, line width=0.4pt}]
\node[font=\large\bfseries, text=black] at (5.5,4.6) {Recognition complexity};
\node[font=\scriptsize, text=black!60] at (5.5,4.15) {(sub-problem \(\times\) grammar class)};
\foreach \x/\lab in {1.1/k\!=\!3, 3.3/k\!=\!2, 5.5/k\!=\!2^+, 7.7/k\!=\!1, 9.9/k\!=\!0} {
  \node[font=\footnotesize\bfseries, text=black] at (\x,3.55) {\(\lab\)};
}
\node[font=\footnotesize\bfseries, text=black, anchor=east, align=right] at (-0.15,2.7) {Membership \\ decision};
\node[font=\footnotesize\bfseries, text=black, anchor=east, align=right] at (-0.15,1.6) {One tree};
\node[font=\footnotesize\bfseries, text=black, anchor=east, align=right] at (-0.15,0.5) {All \\ trees};
\node[hcell, fill=lv0!25, draw=lv0!60] at (1.1,2.7) {\(\Theta(n)\)};
\node[hcell, fill=lv2!40, draw=lv2!70] at (3.3,2.7) {\(O(n^{\omega})\)};
\node[hcell, fill=lv3!40, draw=lv3!70] at (5.5,2.7) {\(O(n^6)\)};
\node[hcell, fill=lv4!40, draw=lv4!70] at (7.7,2.7) {\textsf{PSPACE}-c.};
\node[hcell, fill=lv6!30, draw=lv6!70, text=black] at (9.9,2.7) {undec.};
\node[hcell, fill=lv0!25, draw=lv0!60] at (1.1,1.6) {\(\Theta(n)\)};
\node[hcell, fill=lv2!40, draw=lv2!70] at (3.3,1.6) {\(O(n^{\omega})\)};
\node[hcell, fill=lv3!40, draw=lv3!70] at (5.5,1.6) {\(O(n^6)\)};
\node[hcell, fill=lv4!40, draw=lv4!70] at (7.7,1.6) {\textsf{PSPACE}-c.};
\node[hcell, fill=lv6!30, draw=lv6!70, text=black] at (9.9,1.6) {undec.};
\node[hcell, fill=lv0!25, draw=lv0!60] at (1.1,0.5) {\(O(n)\)};
\node[hcell, fill=lv4b!35, draw=lv4b!70] at (3.3,0.5) {\(O(4^n)\)};
\node[hcell, fill=lv4b!35, draw=lv4b!70] at (5.5,0.5) {\(O(4^n)\)};
\node[hcell, fill=lv5!35, draw=lv5!70, text=black] at (7.7,0.5) {\(O(2^{n^2})\)};
\node[hcell, fill=lv6!30, draw=lv6!70, text=black] at (9.9,0.5) {undec.};
\foreach \y in {3.2, 2.15, 1.05, -0.05} {
  \draw[black!20, line width=0.3pt] (0,\y) -- (11,\y);
}
\end{tikzpicture}
}
\caption{Complexity heatmaps for \textbf{generation} (top) and \textbf{recognition} (bottom). The rows represent the three sub-problems per side; the columns represent the five classes of the Chomsky hierarchy. The palette encodes the order of magnitude: green for linear bounds, yellow for cubic, orange for \(O(n^6)\), red for exponential, dashed dark red for \(O(4^n)\) (Catalan numbers), magenta for \(O(2^{n^2})\), gray for undecidable or unbounded. The \textbf{differential coupling} is visually apparent: the generation matrix (top) varies mainly along the \emph{rows} (sensitivity to the sub-problem: free \(\to\) terminating \(\to\) constrained), whereas the recognition matrix (bottom) varies mainly along the \emph{columns} (sensitivity to the class), with the exception of the ``all trees'' row, which also becomes class-sensitive through combinatorial explosion from \(k=2\) onward.}
\label{fig:heatmaps}
\end{figure}

\textbf{Running example.} For our running musicology example, where the grammar is of level \(k=2^+\), recognizing a polyphonic score of length \(n\) requires in the worst case \(O(n^6)\) operations, whereas generating an arbitrary excerpt requires \(O(n^2)\) to \(O(n^3)\). The gap is of the order of \(n^3\) to \(n^4\): for a 100-bar excerpt (\(n \approx 100\)), this represents a ratio of \(10^6\) to \(10^8\) between the two operations --- significant, but tractable on standard hardware. If one were to promote the same grammar to the context-sensitive level by adding arbitrary global constraints, recognition would become PSPACE-complete, and there would then be no polynomial algorithm to decide that an observed excerpt does belong to the defined language.

\textbf{Counter-argument.} One might object that for CFGs the gap is of the order of \(n^{0.37}\) only if one retains Valiant's bound (\(O(n^{2.37})\)) rather than CYK's (\(O(n^3)\)), and that it even vanishes entirely if one restricts to the LL(\(k\)) or LR(\(k\)) subclasses --- restricted subclasses of CFGs admitting deterministic linear-time parsing --- which parse in \(O(n)\). The computational asymmetry would then reach an order of \(n^3\) to \(n^4\) only from level \(k=2^+\) onward.

\textbf{Resolution.} The counter-argument illustrates the necessity of distinguishing the intrinsic bound from the best-known-algorithm bound. The \(O(n^\omega)\) bound for CFGs is \emph{conditional} on the conjecture \(\omega = 2\): proving that there exists a Boolean matrix-multiplication algorithm in \(O(m^{2+\epsilon})\) for every \(\epsilon > 0\) would indeed make the asymmetry vanish at the CFG level. But this conjecture has resisted every known approach since \citet{strassen1969gaussian}. Under the current state of knowledge, the effective gap for CFGs therefore remains bounded by \(n^{\omega - 1}\). For the LL/LR subclasses, the gap does vanish, which confirms that D1 depends on the precise class considered and not only on the level \(k\) of the hierarchy. In any case, the \(\mathsf{P}\) vs \(\mathsf{PSPACE}\) separation for CSGs and the decidable vs undecidable separation for RE depend on no algorithmic hypothesis: they are intrinsic.

\textbf{Contribution.} Complexity by class of the Chomsky hierarchy has been the subject of much prior work \citep{barton1987complexity}; \citep{hopcroft2006introduction}; \citep{jurafsky2024speech}. Our contribution is not the establishment of new bounds, but the \emph{framing of complexity as a parametric dimension} of the generation-recognition asymmetry, articulated with the five other dimensions (D2 to D6) in the unified taxonomy presented in §4.2 to 4.6. The systematic decomposition of each side into three distinct computational sub-problems --- membership decision, tree construction, enumeration of all trees for recognition; free generation, terminating example-generation, generation under semantic constraint for generation --- makes it possible to locate precisely where the asymmetry is polynomial, where it separates two conjecturally distinct complexity classes, and where it crosses the threshold of decidability. The differential coupling brought to light (recognition sensitive to the class, generation sensitive to the sub-problem) clarifies what remained blurred in introductory presentations, where the six sub-problems are often merged under the single labels ``generation'' and ``recognition''.
\subsection{D2 --- Ambiguity Asymmetry}

\textbf{Thesis.} Generation is a \emph{function}: given \(G\) and a complete \textbf{derivation sequence} \(d\), the output \(w = \text{gen}(G, d)\) is uniquely determined. Parsing is a \emph{relation}: given \(G\) and a string \(w\), the set of valid structural descriptions \(\text{parse}(G, w) = \{t_1, \ldots, t_k\}\) may contain zero, one, or exponentially many elements. The fundamental asymmetry rests on the \textbf{differential informativeness of the input} to the two operations: the generator receives (grammar + complete derivation), an input that fixes the output; the parser receives only (grammar + string), a strictly less informative input that does not suffice to fix the tree. The multivaluedness of parsing therefore reflects the underdetermination of its input, not an intrinsic property of the recognition operation. For certain languages, this multivaluedness is nonetheless \emph{inherent} --- no grammar can eliminate it, even assuming a maximally informative input.

\textbf{Formal argument.} For the generation function:

\[\text{gen}: G \times D \to \Sigma^*, \quad (G, d) \mapsto w\]

where \(D\) is the set of complete derivation sequences. For a fixed \(d\), the output \(w\) is unique: \(\text{gen}\) is a function.

Note that \(\text{gen}\) is not injective in the interesting direction: distinct derivations \(d_1 \neq d_2\) may produce the same string \(w\) (e.g., leftmost and rightmost derivations of the same tree). The generation mapping \(D \to \Sigma^*\) is surjective onto \(L(G)\) but not injective --- multiple derivation paths converge to the same surface form.

For the parsing relation:

\[\text{parse}: G \times \Sigma^* \to 2^{\mathcal{T}}, \quad (G, w) \mapsto \{t_1, t_2, \ldots, t_k\}\]

where \(\mathcal{T}\) is the set of parse trees and \(k = |\text{parse}(G, w)| \geq 0\). When \(k > 1\), the string \(w\) is \emph{ambiguous} with respect to \(G\). The growth of \(k\) is governed by the Catalan numbers:

\[C_n = \frac{1}{n+1}\binom{2n}{n} \sim \frac{4^n}{n^{3/2}\sqrt{\pi}}\]

A sentence with 4 prepositional phrases admits \(C_4 = 14\) distinct attachment trees \citep{church1982coping} (with 3 phrases, the count is \(C_3 = 5\)). Shared parse forests (\citealt{billot1989shared}) provide compact \(\mathcal{O}(n^3)\) representations of exponentially many trees, but the underlying multivaluedness is irreducible.

\textbf{Running example.} The generator of \emph{``I saw the man with the telescope''} has a single derivation in mind --- say, VP-attachment (the speaker uses the telescope). Generation is a function: one derivation \(d\), one string \(w\). The parser, receiving only \(w\), must produce both parse trees: \(t_1\) (VP-attachment: \emph{saw-with-telescope}) and \(t_2\) (NP-attachment: \emph{man-with-telescope}). It cannot determine which the speaker intended without extra-grammatical information --- the ambiguity is structural and irreducible at the syntactic level.

\textbf{Inherent ambiguity.} The multivaluedness of parsing is not merely a flaw of particular grammars but a \emph{property of formal languages}.

\textbf{Theorem} (\citealt{parikh1966context}). There exist context-free languages \(L\) such that every CFG \(G\) with \(L(G) = L\) is ambiguous. Such languages are called \emph{inherently ambiguous}.

The canonical example:

\[L = \{a^n b^n c^m d^m \mid n, m \geq 1\} \cup \{a^n b^m c^m d^n \mid n, m \geq 1\}\]

\textbf{Theorem} (\citealt{brabrand2007ambiguity}). It is undecidable whether an arbitrary CFG \(G\) is ambiguous.

These results establish that ambiguity cannot, in general, be designed away. One cannot even check, in general, whether a given grammar exhibits it. The parsing relation is irreducibly multivalued for any sufficiently expressive grammar.

\textbf{Semiotic dimension.} Eco's \emph{Opera Aperta} \citeyearpar{eco1962opera} anticipated this observation in aesthetic terms: the ``open work'' admits multiple valid interpretations --- a literary formulation of the same structural property. A single production (the work) admits multiple analyses; the generative act is convergent (many intentions \(\to\) one work) while the analytical act is divergent (one work \(\to\) many readings). This cardinality asymmetry is not specific to formal grammars --- it is inherent in any expressive system where meaning is generated by one agent and interpreted by another.

\textbf{Counter-argument.} Deterministic context-free languages (DCFL, recognized by a deterministic pushdown automaton) are always unambiguous \citep[Thm~6.20]{hopcroft2001introduction}. Does this not show that ambiguity is avoidable?

\textbf{Resolution.} DCFLs are a proper subset of all CFLs and exclude all natural languages. Moreover, \citet{parikh1966context} proved that \emph{inherently ambiguous} context-free languages exist --- languages for which no unambiguous grammar can be written. The ambiguity of parsing is a property of languages, not a defect of grammars. As \citet{delahiguera2010inference} summarizes: ``All reasonable questions relating to ambiguity are undecidable.''

\subsection{D3 --- Directionality Asymmetry}

\textbf{Thesis.} For phrase-structure grammars, two directions must be distinguished from the outset. The \textbf{semantic direction} (from intention to surface: from meaning to the produced string) is tautologically \emph{top-down} for any generative operation, by the very definition of what it is to generate --- starting from a high-level intention and materializing it on the surface. The \textbf{syntactic direction} (from the axiom to the terminal symbols, in the derivation), by contrast, is fixed for generation but admits several algorithmic choices for parsing: top-down, bottom-up, or hybrid (Earley, left-corner). The central content of D3 is this \textbf{directional degree of freedom} that parsing possesses and generation does not: parsing offers a range of algorithmic strategies for reconstructing the structure, whereas generation has no analogue to this choice --- the semantic direction being tautologically constrained, and the syntactic direction being fixed by the very nature of derivation. For non-phrase-structure formalisms (L-systems, shape grammars, emergent generation), the generative direction remains fixed but takes formalism-specific forms --- temporal rather than hierarchical for L-systems, geometric for shape grammars.

\textbf{Formal argument.} Every derivation in a phrase-structure grammar proceeds:

\[S \Rightarrow \alpha_1 \Rightarrow \alpha_2 \Rightarrow \cdots \Rightarrow w \in \Sigma^*\]

This process is \emph{inherently top-down}: it moves from the most abstract level (\(S\), the intention: ``I want a musical phrase'') to the most concrete (\(w\), the surface: the actual notes). The structure is vertical (hierarchical), not horizontal (sequential) --- the temporal left-to-right order of the output is a by-product of the hierarchical decomposition \(S \to AB, A \to \text{do re}, B \to \text{mi fa}\).

\textbf{Scope.} D3 as stated applies to \emph{phrase-structure grammars} (the Chomsky hierarchy: Types 0--3), where derivation proceeds by replacing nonterminals in sentential forms. Alternative formalisms require separate analysis:

\textbf{L-systems} (\citealt{lindenmayer1968mathematical}) employ parallel rewriting: all symbols are rewritten simultaneously at each step. There is no axiom-to-terminal direction in the phrase-structure sense --- the ``generation'' is iterative, not hierarchical. However, even in L-systems, the developmental direction is fixed (step \(t \to\) step \(t+1\)), while analysis (inferring the L-system from an observed structure) requires inverse reasoning. The asymmetry persists but takes a different form: \emph{temporal} rather than \emph{hierarchical}.

\textbf{Shape grammars} (\citealt{stiny1972shape}) operate on geometric structures where ``top-down'' has no clear meaning. Yet the generative direction (apply rules to produce shapes) remains fixed, while the analytical direction (infer which rules produced an observed shape) faces the same combinatorial explosion as parsing.

\textbf{Emergent generation} (cellular automata, multi-agent systems) produces structure bottom-up without an explicit grammar. These systems \textbf{sit at the boundary of our framework}: their local rules can be formalized as parallel rewriting systems (cf.~L-systems, §4.3 D3), but the generated language is not always defined in the classical sense.

One apparent counter-example within phrase-structure grammars: \citet{kay1996chart} proposed a chart-based generation algorithm where the syntactic assembly may proceed bottom-up. But even there, the \emph{semantic} direction remains top-down --- what drives the process is still intention → surface. The mechanism may assemble pieces in any order; the logic always flows from abstract to concrete.

For parsing, the direction is a \emph{design parameter} \(\delta\):

\[\text{parse}_\delta(G, w) = \{t \in \mathcal{T} \mid \text{yield}(t) = w\}\]

where \(\delta\) denotes the parsing strategy, chosen from the following:

\begin{longtable}[]{@{}
  >{\raggedright\arraybackslash}p{(\columnwidth - 4\tabcolsep) * \real{0.3}}
  >{\raggedright\arraybackslash}p{(\columnwidth - 4\tabcolsep) * \real{0.2}}
  >{\raggedright\arraybackslash}p{(\columnwidth - 4\tabcolsep) * \real{0.5}}@{}}
\toprule\noalign{}
\begin{minipage}[b]{\linewidth}\raggedright
Strategy
\end{minipage} & \begin{minipage}[b]{\linewidth}\raggedright
Direction
\end{minipage} & \begin{minipage}[b]{\linewidth}\raggedright
Mechanism
\end{minipage} \\
\midrule\noalign{}
\endhead
\bottomrule\noalign{}
\endlastfoot
LL (recursive descent) & Top-down & Predict nonterminal, match terminal \\
LR (shift-reduce) & Bottom-up & Shift terminals, reduce to nonterminals \\
Earley (1970) & Hybrid & Top-down prediction + bottom-up completion \\
CYK & Bottom-up & Dynamic programming on substrings \\
Left-corner (\citealt{rosenkrantz1970deterministic}) & Hybrid & Bottom-up initiation + top-down prediction \\
\end{longtable}

The Dragon Book (\citealt{aho1986compilers}) devotes entire chapters to this choice --- a choice that exists \emph{only for the parser}. The generator has no analogous design decision.

\textbf{Running example.} The generator of \emph{``I saw the man with the telescope''} proceeds top-down: \(S \Rightarrow \text{NP}\ \text{VP} \Rightarrow \text{I}\ \text{VP} \Rightarrow \text{I}\ \text{V}\ \text{NP}\ \text{PP} \Rightarrow \cdots\) --- the direction is fixed by the derivation. The parser, faced with the same string, can proceed bottom-up (shift \emph{``I''}, shift \emph{``saw''}, reduce to V, \ldots), top-down (predict \(S\), predict NP, match \emph{``I''}, \ldots), or hybrid (Earley). Each strategy handles the PP-attachment ambiguity differently: a bottom-up parser discovers both attachment sites during reduction; a top-down parser must predict both before matching. The generator makes no such choice.

\citet{shieber1988uniform} provides concrete evidence. In parsing, the chart is indexed by \emph{string positions} --- ``from position 3 to position 7, a noun phrase has been recognized.'' In generation, \emph{the string does not yet exist} --- it is the output being constructed. Chart-based generation therefore cannot reuse the parser's indexing scheme. \citet{shieber1990semantic} resolved this by driving generation from \emph{semantic heads} rather than syntactic positions --- the ``semantic head-driven generation'' algorithm. This restructuring is not a parameter change: it is a fundamental reversal of the data flow. Parsing goes from surface to structure (\(w \to t\)); generation goes from structure to surface (\(\phi \to w\)). What serves as input to one operation does not yet exist for the other. The same declarative grammar requires two distinct procedural interpretations.

Knuth's \citeyearpar{knuth1968semantics} attribute grammars formalize the same asymmetry in the semantic domain. \emph{Inherited attributes} flow top-down (\(\downarrow\), parent to child), while \emph{synthesized attributes} flow bottom-up (\(\uparrow\), child to parent). \citet[§15.3]{grune2008parsing} note that, although defined as generative devices, attribute grammars are mainly used for recognition tasks (semantic checking, type inference, code generation from parse trees); their bidirectional attribute flow is an \emph{analytical} property.

\textbf{Counter-argument.} The NLG pipeline \citep{reiter2000building} could be replaced by integrated architectures (KAMP, neural end-to-end). Does the top-down direction always hold?

\textbf{Resolution.} The \emph{syntactic} construction can vary (bottom-up composition in CCG, for instance). But the \emph{semantic} direction --- from intention to surface, from abstract to concrete --- remains top-down. The pipeline can be compressed, but the direction of meaning flow cannot be reversed. D3 is a claim about the \emph{logical order} of the generative process, not about its implementation.

\textbf{Contribution.} Although the directional aspects of parsing (top-down vs bottom-up, LL vs LR) and the practical asymmetry between parsing and generation algorithms have been studied extensively (Russell, Carroll \& Warwick 1990; Shieber 1988; Strzalkowski 1993), the explicit framing of directionality as a structural dimension within a unified taxonomy of the generation-recognition asymmetry has not, to our knowledge, been articulated previously.

\subsection{D4 --- Information Asymmetry}

\textbf{Thesis.} The generator has access to all the source information --- intention, context, constraints, encyclopedic knowledge. The recognizer has access only to the \textbf{observable surface}: the internal structure (derivation tree) remains reconstructible with the grammar, but the \textbf{extra-grammatical information} that guided the choice of that structure among the possible alternatives is lost in the channel and must be inferred. The gap is structural (Shannon, through the \textbf{non-injectivity of the intention \(\to\) string encoding}) and intentional (Eco, through the \textbf{narcotization of properties}).

\textbf{Formal argument.} In Shannon's framework, the encoder (generator) knows the source message \(X\) with certainty:

\[H(X \mid X) = 0\]

The decoder (recognizer) receives the signal \(Y\) through a noisy channel and must infer \(X\), suffering the equivocation \(H(X \mid Y) > 0\). The mutual information quantifies what survives:

\[I(X; Y) = H(X) - H(X \mid Y)\]

The \emph{noise} in the grammatical analogy is the \textbf{extra-grammatical information lost in the channel}: the speaker's intention, the pragmatic context, the shared encyclopedic knowledge. The recognizer must reconstruct this information by inference --- a non-triviality. The generator operates by \emph{deduction} (known premises → conclusions); the parser operates by \emph{abduction} (observed effects → probable causes).

\citet{eco1979lector} provides the semiotic complement: the text is a \emph{meccanismo pigro} (lazy mechanism) that ``lives on the surplus of meaning introduced by the addressee.'' The generator deliberately omits information through the \emph{narcotization} of properties, trusting the decoder to fill the gaps. This is not a flaw but a \emph{design feature}: expressive systems are asymmetric by construction.

\textbf{Running example.} The speaker of \emph{``I saw the man with the telescope''} knows the full pragmatic context: they were at an observatory, using a telescope to observe someone. This contextual information (\(H(X \mid X) = 0\)) disambiguates the PP-attachment completely. But the listener receives only the seven-word string --- the linearized surface. The hierarchical structure (which PP attaches where) has been flattened into a sequence, and the pragmatic context that would disambiguate it has been lost in the channel. The listener must reconstruct it by abduction: ``given this string, what is the most probable intended structure?''

\textbf{Counter-argument.} The Bits-Back argument (\citealt{frey1996free}) shows that the decoder's uncertainty can be exploited as an auxiliary information channel --- the asymmetry becomes a \emph{resource}.

\textbf{Resolution.} The Bits-Back argument does not eliminate D4; it reframes it. The generator still knows \(X\) with certainty while the parser does not. What changes is the \emph{valuation} of the gap: from a pure deficit to a potential resource. This enriches D4 without undermining it.

\textbf{Note on the D4/D6 independence.} The psycholinguistic \emph{P-chain} framework \citep{dell2014pchain} proposes that the production system implements implicit prediction during comprehension. Under this thesis, D4 (static information gap) and D6 (dynamic temporal gap) may not constitute two separate cognitive mechanisms, contrary to what we claim in §4.6 below --- the static gap D4 could be the aggregate of the accumulated incremental surprises (D6). Our framework remains defensible at the level of formal language theory (where the two processes can be analyzed as distinct computational objects), but the cognitive question of their mechanistic separation remains open.

\subsection{D5 --- Grammar Inference Asymmetry}

\textbf{Thesis.} Beyond the generation-recognition duality, grammar inference constitutes a third operation that is not a separate field but the \emph{extreme case} of recognition under decreasing grammatical knowledge. Between the extremes of full grammatical knowledge (recognition) and no knowledge (inference from scratch) lies a continuum of partial knowledge:

\begin{itemize}

\item
  \textbf{Single known grammar}: standard recognition (§4.1).
\item
  \textbf{Set of candidate grammars}: language identification --- recognition with model selection.
\item
  \textbf{Partially inferred grammar}: the analyst has fragments, built from preprocessing, previous analyses, or during the current parse. Angluin's L* \citeyearpar{angluin1987learning} explicitly occupies this middle ground --- inference that embeds a recognition oracle.
\item
  \textbf{Tabula rasa}: inference from positive data alone. \citet{gold1967language} establishes that this is impossible for superfinite classes.
\end{itemize}

Gold's impossibility result marks a \textbf{qualitative threshold} within this continuum: any finite prior knowledge makes learning feasible in principle (probabilistic Bayesian framework, \citealt[ch.~16]{delahiguera2010inference}; see also \citealt[§3.6.2]{delahiguera2005bibliographical}), while zero prior knowledge makes it impossible for superfinite classes. The transition from ``some knowledge'' to ``no knowledge'' is a phase transition, not a gradual degradation.

A formal characterization of this knowledge gradient, and of the relation between incremental prediction and learning, has been explored in psycholinguistics under the \emph{P-chain} framework \citep{dell2014pchain,gambi2017models}, where ``prediction error drives learning.'' Our formal-language-theory framework here addresses the asymmetry at a different level of abstraction; a unified treatment integrating the two perspectives is beyond the scope of this study.

\textbf{Formal argument.} \citet{gold1967language} proved the fundamental impossibility result:

\textbf{Theorem} \citep{gold1967language}. Let \(\mathcal{L}\) be a class of languages that contains all finite languages and at least one infinite language (a \emph{superfinite} class). Then \(\mathcal{L}\) is not identifiable in the limit from positive data alone --- no algorithm can guarantee convergence to the correct grammar by seeing only valid examples.

For a fixed grammar class and the simplest task on each side, the three operations form a hierarchy of increasing difficulty. Taking the canonical case of context-free grammars (\(k = 2\)), free generation, and membership testing:

\[\underbrace{\text{Generation}}_{\text{free: } \mathcal{O}(n)} \;<\; \underbrace{\text{Recognition}}_{\text{membership: } \mathcal{O}(n^3)} \;<\; \underbrace{\text{Inference}}_{2^{\mathcal{O}(n)}}\]

As §4.1 established, however, both generation and recognition span wide complexity ranges depending on task specification and grammar class --- from \(\mathcal{O}(n)\) to \(\notin \mathsf{R}\). The ordering above captures the asymmetry for the canonical case; whether it holds uniformly across the full parameter space of §4.1, and how the three-way gap evolves as both axes vary, remains an open research problem requiring dedicated formal investigation.

The inference bound reflects the NP-hardness of minimum-consistent-DFA identification (Gold, 1978; \citealt{pitt1993minimum}). More precisely, \citet{pitt1993minimum} showed that the minimum-consistent-DFA problem is not approximable within any polynomial --- a result strictly stronger than NP-hardness. \citet{kearns1994cryptographic} further showed that PAC-learning DFAs is as hard as breaking RSA encryption --- establishing a cryptographic lower bound on the difficulty of inference.

\begin{figure}[H]
\centering
\resizebox{\columnwidth}{!}{%
% Figure 2 — The Three Operations: A Hierarchy of Difficulty
\begin{tikzpicture}[scale=1.5, transform shape, every node/.style={font=\small, text=black}]
  % Triangle
  \node[draw=green!50!black, rounded corners, fill=green!8, text=black, font=\normalsize\bfseries, minimum width=3cm, minimum height=1cm, align=center]
    (gen) at (0,0) {Generation \\ \(\mathcal{O}(n)\)};
  \node[draw=yellow!50!black, rounded corners, fill=yellow!8, text=black, font=\normalsize\bfseries, minimum width=3cm, minimum height=1cm, align=center]
    (rec) at (0,2.5) {Recognition \\ \(\mathcal{O}(n^3)\)};
  \node[draw=red!50, rounded corners, fill=red!8, text=black, font=\normalsize\bfseries, minimum width=3cm, minimum height=1cm, align=center]
    (inf) at (0,5) {Inference \\ \(2^{\mathcal{O}(n)}\)};
  % Arrows with labels
  \draw[->, thick, black!50] (gen) -- (rec) node[midway, right, font=\footnotesize, text=black!60] {given \(G\), find structure of \(w\)};
  \draw[->, thick, black!50] (rec) -- (inf) node[midway, right, font=\footnotesize, text=black!60] {given \(\{w_i\}\), find \(G\)};
  % Left labels
  \node[left=0.8cm, font=\footnotesize, text=black!60, align=left, text width=3.5cm] at (gen.west) {Given: \(G\) \\ Sought: \(w\)};
  \node[left=0.8cm, font=\footnotesize, text=black!60, align=left, text width=3.5cm] at (rec.west) {Given: \(G, w\) \\ Sought: structure};
  \node[left=0.8cm, font=\footnotesize, text=black!60, align=left, text width=3.5cm] at (inf.west) {Given: \(\{w_1, \ldots, w_k\}\) \\ Sought: \(G\)};
  % Difficulty arrow
  \draw[->, very thick, black!50] (6,0) -- (6,5) node[midway, right, font=\small, text=black!50] {increasing difficulty};
\end{tikzpicture}
}
\caption{The three operations: a hierarchy of difficulty. \emph{Canonical case ($k = 2$, simplest tasks). Full ranges: $\mathcal{O}(n)$ to $\notin \mathsf{R}$ — see §4.1.}}
\label{fig:hierarchy}
\end{figure}

The inequality between recognition and inference is more pronounced than that between generation and recognition: \citet{kearns1994cryptographic} reduced PAC-learning of DFAs to cryptographic problems, showing that an efficient PAC learner for DFAs would imply efficient algorithms for inverting RSA encryption \citep[§9.4]{delahiguera2010inference}. Inference is not merely harder --- it is \emph{qualitatively} harder.

\textbf{Running example.} Given our sentence \emph{``I saw the man with the telescope''} and a known English grammar \(G\), \emph{recognition} produces two parse trees --- hard but feasible. Now imagine receiving this single sentence with \emph{no grammar at all}: can you infer the rules of English? Even determining whether ``with'' introduces a prepositional phrase, an instrumental adjunct, or a comitative marker requires prior linguistic knowledge. A single sentence is radically underdetermined --- Gold's theorem in miniature. Inference occupies a different regime entirely.

\citet{delahiguera2010inference} situates grammar inference as the converse of generation: where generation applies grammatical rules to produce strings, inference must recover the rules from the strings. This connects D5 directly to the asymmetry. The compression perspective reinforces this: algorithmic information theory (\citealt{chaitin2005metamath}; see also \citealt{rissanen1978modeling} and \citealt{grunwald2007mdl} for the formal MDL principle) frames scientific understanding itself as an act of compression. Chaitin states this principle explicitly: ``A theory is good to the extent that it compresses the data into a much smaller set of theoretical assumptions. The greater the compression, the better!'' (\citealt{chaitin2005metamath}). Inference thus becomes equivalent to finding the shortest description of the observed data (the MDL principle). Generation = decompression (expanding a compact description); inference = compression (finding that description). The asymmetry between compression (hard) and decompression (easy) is another manifestation of the same divide.

\textbf{Counter-argument (C4).} PCFGs escape Gold's theorem: learning stochastic context-free grammars from positive data in a Bayesian framework is possible in principle via the probabilistic prior that breaks Gold's symmetry (\citealt[ch.~16]{delahiguera2010inference}; \citealt[§3.6.2]{delahiguera2005bibliographical} for a review of the Baker 1979, Lari \& Young 1990, and successor results). Angluin's L* algorithm (1987) learns regular languages efficiently with a membership oracle.

\textbf{Resolution.} These results restrict but do not eliminate D5. PCFGs shift the problem from ``impossible'' to ``extremely difficult in practice'' --- Kearns and Valiant's (1994) RSA reduction still applies to deterministic classes. L* requires a \emph{membership oracle} --- embedding a recognition capability within the inference process. The hierarchy gen \textless{} recog \textless{} inference holds for fixed grammar class and minimal tasks, but the exact gap depends on the framework (deterministic = impossible, probabilistic = very hard, with oracle = feasible but presupposes recognition). A comprehensive formal treatment --- integrating the two-dimensional complexity landscape of §4.1 with the knowledge gradient introduced above --- is needed to fully characterize how the three-way asymmetry evolves across grammar classes and task specifications.

\subsection{D6 --- Temporal Asymmetry}

\textbf{Thesis.} Generation is \emph{causal}: the system creates the future, with zero uncertainty about forthcoming symbols. Parsing is \emph{expectation-based} \citep{levy2008expectation}: the system must infer structure from a sequence that unfolds incrementally, updating its predictions as each new token arrives. The \textbf{surprisal theory} \citep{hale2001probabilistic,levy2008expectation} formalizes this asymmetry but has never been framed as such.

\textbf{Formal argument.} Define the \emph{surprisal} of the \(i\)-th token given its preceding context:

\[S(w_i) = -\log_2 P(w_i \mid w_1, w_2, \ldots, w_{i-1})\]

For a \emph{deterministic generator}, the system knows which token it will produce at each step:

\[P_{\text{gen}}(w_i \mid w_1, \ldots, w_{i-1}) = 1 \quad \Longrightarrow \quad S_{\text{gen}}(w_i) = 0\]

For an \emph{incremental parser}, the system has observed \(w_1, \ldots, w_i\) but not \(w_{i+1}, \ldots, w_n\). It maintains competing hypotheses about the derivation:

\[P_{\text{parse}}(w_i \mid w_1, \ldots, w_{i-1}) < 1 \quad \Longrightarrow \quad S_{\text{parse}}(w_i) > 0\]

\textbf{The surprisal quantifies the temporal asymmetry}: exactly zero for the generator, strictly positive for the parser.

\textbf{Running example.} Consider incremental parsing of \emph{``I saw the man with the telescope.''} After processing \emph{``I saw the man,''} the parser maintains a single dominant hypothesis (simple transitive sentence). The word \emph{``with''} triggers a surprisal spike: two competing attachment sites suddenly become available, and the parser must split its probability mass between VP-attachment and NP-attachment. The subsequent words \emph{``the telescope''} may partially disambiguate (telescopes are instruments, favoring VP-attachment) but the structural uncertainty persists until the sentence boundary. The generator, by contrast, experiences \(S = 0\) at every position --- including at \emph{``with,''} where the speaker already knows the intended attachment.

\textbf{Independence from D4.} Consider a non-incremental parser with full lookahead --- one that receives the complete string \(w = w_1 \cdots w_n\) before beginning analysis. Such a parser suffers D4 (it has only the surface, not the intention) but does \textbf{not} suffer D6: there is no temporal unfolding, no token-by-token surprisal, no expectation-based processing. Conversely, a stochastic generator with a temperature parameter suffers a form of D6 (non-zero surprisal at each step) but does \textbf{not} suffer D4 (it has full access to its own model parameters).

D4 is a \emph{static} gap: what each agent knows in total. D6 is a \emph{dynamic} gap: how uncertainty evolves over the sequential processing of the string. A batch parser (CYK) suffers D4 but not D6. An incremental parser (Earley prefix probabilities) suffers both. This confirms their independence at the formal level --- the psycholinguistic \emph{prediction-by-production} perspective challenges this independence at the level of cognitive mechanisms (see the Acknowledgment below).

\begin{figure}[H]
\centering
\resizebox{\columnwidth}{!}{%
% Figure 3 — Surprisal on running example: Generator vs. Parser
\begin{tikzpicture}[scale=1.5, transform shape]
  % Axes
  \draw[black!50, line width=0.6pt] (0,0) -- (10.5,0);
  \draw[black!50, line width=0.6pt] (0,0) -- (0,4.2);
  \node[font=\small, text=black!60, above] at (0,4.2) {\(S(w_i)\)};
  % Token labels on x-axis
  \foreach \x/\w in {1/I, 2.3/saw, 3.6/the, 4.9/man, 6.2/with, 7.5/the, 8.8/telescope} {
    \draw[black!30] (\x,0) -- (\x,-0.1);
    \node[font=\footnotesize, text=black, below] at (\x,-0.1) {\textit{\w}};
  }
  % Y-axis ticks
  \foreach \y/\lab in {1/low, 2/med, 3/high} {
    \draw[black!20] (0,\y) -- (10.2,\y);
  }
  % Generator: flat at 0 (blue line)
  \draw[blue!70!black, line width=2pt] (1,0) -- (2.3,0) -- (3.6,0) -- (4.9,0) -- (6.2,0) -- (7.5,0) -- (8.8,0);
  \foreach \x in {1, 2.3, 3.6, 4.9, 6.2, 7.5, 8.8} {
    \fill[blue!70!black] (\x,0) circle (2.5pt);
  }
  \node[font=\footnotesize\bfseries, text=blue!70!black] at (5,0.4) {Generator: \(S = 0\)};
  % Parser: surprisal profile (red line)
  % I=0.8 saw=1.5 the=0.6 man=1.0 with=3.4(spike) the=1.8 telescope=1.2
  \fill[red!8, opacity=0.5] (1,0) -- (1,0.8) -- (2.3,1.5) -- (3.6,0.6) -- (4.9,1.0)
    -- (6.2,3.4) -- (7.5,1.8) -- (8.8,1.2) -- (8.8,0) -- cycle;
  \draw[red!70!black, line width=2pt] (1,0.8) -- (2.3,1.5) -- (3.6,0.6) -- (4.9,1.0)
    -- (6.2,3.4) -- (7.5,1.8) -- (8.8,1.2);
  \foreach \x/\y in {1/0.8, 2.3/1.5, 3.6/0.6, 4.9/1.0, 6.2/3.4, 7.5/1.8, 8.8/1.2} {
    \fill[red!70!black] (\x,\y) circle (2.5pt);
  }
  \node[font=\footnotesize\bfseries, text=red!70!black] at (7.8,3.6) {Parser: \(S > 0\)};
  % Annotation: spike at "with"
  \draw[black!40, thin, dashed] (6.2,0) -- (6.2,3.4);
  \node[font=\scriptsize\itshape, text=black!50, anchor=south west] at (6.35,3.0) {PP-attachment};
  \node[font=\scriptsize\itshape, text=black!50, anchor=south west] at (6.35,2.6) {ambiguity};
\end{tikzpicture}
}
\caption{Surprisal asymmetry on the running example \emph{"I saw the man with the telescope."} The generator (blue) experiences $S = 0$ at every position. The parser (red) shows a surprisal spike at \emph{"with"} — the PP-attachment ambiguity point. The shaded area represents the cumulative temporal asymmetry.}
\label{fig:surprisal}
\end{figure}

\citet{hale2001probabilistic} proposed surprisal as a psycholinguistic model: processing difficulty is proportional to \(S(w_i)\). \citet{levy2008expectation} refined this into \emph{expectation-based comprehension}: difficulty is the cost of reallocating probability mass across competing hypotheses when a new word arrives. \citet{stolcke1995efficient} provided computational machinery: a probabilistic Earley parser computing \emph{prefix probabilities} \(P(w_1 \cdots w_i)\) at each position.

The asymmetry manifests operationally in \emph{lookahead}. An LL(\(k\)) parser requires \(k\) tokens of lookahead to resolve decisions --- a necessity with no analog in generation. The generator creates the future; it needs no preview. \citet{nivre2008algorithms} showed that incremental parsing can achieve \(\mathcal{O}(n)\) --- but only by sacrificing optimality (greedy choices). To match the generator's efficiency, the parser must abandon the guarantee of finding the best parse tree.

\textbf{Connection to signal processing.} The temporal asymmetry has a precise analog in DSP: \emph{causal} vs.~\emph{anti-causal} systems. A causal system's output at time \(t\) depends only on inputs at times \(\leq t\) (like a generator); an anti-causal system can access future inputs (like a parser with full lookahead). This analogy is well-established in the signal processing literature. \citet{dubnov2024deep} draw an analogous distinction in audio: causal models (autoregressive, e.g.\ GPT) generate sample by sample from the past context alone, whereas bidirectional models (e.g.\ BERT) access the complete sequence and operate as analyzers rather than generators. The causal attention mask in autoregressive Transformers \citep{vaswani2017attention} imposes this asymmetry by hiding future positions during training and inference.

The historical record provides a striking perspective: human beings have produced sound --- generation --- since prehistory (bone flutes dating to 27,000 BCE; \citealt{roederer2008physics}), yet formal analysis of sound came millennia later, with Mersenne's \emph{Harmonie Universelle} (1636) as the earliest systematic treatment. The generation-recognition asymmetry predates formal language theory by centuries.

\textbf{Counter-argument.} Stochastic generation (sampling from a language model) involves uncertainty: \(P(w_i) < 1\) before sampling, so \(S_{\text{gen}} > 0\).

\textbf{Resolution.} In stochastic generation, the uncertainty is \emph{chosen} (the generator controls the temperature, the sampling strategy). In parsing, the uncertainty is \emph{imposed} by the input. The distinction between chosen and imposed uncertainty is the core of D6.

\textbf{Acknowledgment of an alternative perspective (P-chain).} An alternative perspective comes from psycholinguistics: \citet{dell2014pchain} propose that the production system itself implements implicit prediction during comprehension (the \emph{prediction-by-production} hypothesis, see also \citealt{martin2018prediction}; \citealt{gambi2017models}; \citealt{gastaldon2024predictive}). Under this view, comprehension uses the generator implicitly, partly blurring the temporal asymmetry described here. Our framework remains useful at the level of formal language theory --- where the two processes can be analyzed as distinct computational objects --- but the cognitive question of their mechanistic separation remains open.

\textbf{Contribution.} Surprisal theory is among the most cited frameworks in computational psycholinguistics (Hale 2001 and Levy 2008 together accumulating more than 3{,}000 citations according to Semantic Scholar). Although it was developed in psycholinguistics with an implicit perspective of connection to the production-comprehension asymmetry (cf.\ P-chain), it has not, to our knowledge, been explicitly framed as formalizing the \emph{temporal dimension} of a unified taxonomy of the generation-recognition asymmetry in formal language theory. This paper proposes that connection at this level of abstraction.

\section{Case Studies}

The six dimensions are not merely theoretical constructs. This section examines four domains where the asymmetry manifests concretely, illustrating how the dimensions interact in practice.

\subsection{The compiler asymmetry: LL vs.~LR}\label{the-compiler-asymmetry-ll-vs.-lr}

The Dragon Book (\citealt{aho1986compilers}) provides an unintentional but compelling case study. Its treatment of parsing dwarfs its treatment of code generation --- hundreds of pages of parsing theory versus a relatively contained discussion of tree-rewriting and pattern matching.

Within parsing, the LL-versus-LR distinction embodies D3. LL parsing \emph{imitates} the generative process: top-down, predicting which production to apply. LR parsing \emph{inverts} it: bottom-up, shifting terminals and reducing them to nonterminals. Crucially, LR is \emph{more powerful} --- the class of LR(\(k\)) languages properly includes LL(\(k\)) for all \(k\) \citep{knuth1965translation}.

This has a striking interpretation: the parsing strategy that is \emph{most different from generation} (bottom-up, inverse) is the most powerful. The strategy that \emph{imitates generation} (top-down, predictive) is weaker. In Appel's formulation, LR defers its decisions --- it accumulates evidence before committing. This deferral is a capability that only the parser needs; the generator commits immediately.

The compiler case also illustrates D1: code generation is typically \(\mathcal{O}(n)\) (linear AST traversal), while parsing is \(\mathcal{O}(n^3)\) for general CFGs. The asymmetry is so familiar to compiler designers that it has become invisible --- which is precisely why D1-D3 have not been identified as dimensions of a unified phenomenon.

\subsection{The NLP asymmetry: from Appelt to Strzalkowski}

The NLP community has engaged with the asymmetry more directly than any other field, yet without treating it as a unified topic.

\citet{appelt1985planning,appelt1987bidirectional} articulated the central tension: ``the most fundamental requirement of any bidirectional grammar is that it be represented \emph{declaratively}. If procedural, asymmetry is inevitable.'' His KAMP system attempted to unify planning (generation) and parsing in a single architecture. The result was computationally impractical --- a cautionary demonstration that bridging the asymmetry is possible in principle but expensive in practice.

\citet{strzalkowski1993reversible,strzalkowski1994general} pursued \emph{grammar inversion}: automatically transforming a parsing grammar into a generation grammar by inverting the control structure while preserving the declarative content. The inverted grammars worked but were less efficient, and certain constructions resisted inversion entirely --- a concrete manifestation of D3 (direction cannot always be reversed mechanically).

\citet{su2019dual} formalized the NLU-NLG relationship as a ``dual problem pair'' with shared latent variables --- the closest the NLP literature has come to treating the asymmetry structurally. Their follow-up (\citealt{su2020towards}) extends this to unsupervised settings. These frameworks stop short of identifying the multiple independent dimensions we analyze here.

\citet{goodman2009generation} provided empirical evidence: testing a broad-coverage HPSG grammar bidirectionally revealed errors invisible to unidirectional use, increasing coverage by 18\%. This directly supports our thesis: the asymmetry has practical consequences for grammar quality.

\subsection{The transformation paradox}

\citet{birman1973parsing} introduced top-down parsing schemata (TS) as alternatives to Chomsky grammars for specifying parsers. Their work reveals an apparent paradox: whereas transforming a generative grammar into a parser requires substantial effort (adding lookahead, disambiguation, and error recovery), transforming a parsing schema into a generator is comparatively straightforward --- one need only execute the derivation paths that the schema has already identified.

At first glance, this \emph{inverts} the asymmetry. The resolution lies in distinguishing \emph{transforming between modes} from \emph{operating within a mode}.

A parsing schema already contains the structural analysis: it has resolved ambiguities, explored the derivation space, encoded the recognition strategy. Converting this into a generator is an ``informational downgrade'' --- one discards the analytical structure and keeps only the generative path. Converting a generative grammar into a parser is an ``informational upgrade'' --- one must \emph{add} the analytical machinery (lookahead, disambiguation, error recovery) absent from the grammar.

This observation \emph{confirms} the asymmetry: parsing contains more information than generation. The asymmetry of \emph{transformation} is the mirror image of the asymmetry of \emph{operation} --- because the parser has already paid the cost that the generator avoids.

\subsection{Grammar inference: from Gold to PCFG}

The trajectory from \citet{gold1967language} to modern inference illustrates D5 in action.

Gold's impossibility result established the theoretical floor: no superfinite class is identifiable from positive data. \citet{horning1969study} provided the first escape: PCFGs are learnable in a Bayesian framework because the probabilistic prior breaks the symmetry that Gold exploits. But the computational cost remains prohibitive for large grammars.

Angluin's L* \citeyearpar{angluin1987learning} showed that regular languages can be learned efficiently --- but requires a \emph{membership oracle}: a recognition capability embedded within the inference process. Inference, even when feasible, presupposes recognition.

The MDL approach (\citealt{rissanen1978modeling}; \citealt{grunwald2007mdl}) reframes inference as compression: the best grammar is the shortest total description of itself plus the data. This makes the connection to the asymmetry explicit: inference finds the \emph{most compact generator}; recognition verifies that the generator accounts for the data; generation unfolds the compact description. The three operations form a cycle, with inference as the most expensive step.

Recent work by \citet{kanani2023grammar} combines Sequitur grammar induction with grammar mutations, and \citet{tsushima2017generative} use PCFG induction with latent variables. These results demonstrate that inference remains an active frontier --- four decades after Gold, the field still seeks practical solutions to what is, in our framework, the \emph{third and hardest direction} of grammar use.

\section{Discussion}

\subsection{The asymmetry is structural, not accidental}

The six dimensions share a common feature: each is independent of the others, and none can be reduced to an implementation detail or a consequence of technological limitations.

D1 is a consequence of the Chomsky hierarchy and the cross-product of task specifications --- it will not change with faster hardware. D2 follows from the mathematical properties of the parsing relation --- no algorithm can make an inherently ambiguous language unambiguous, and the set-valuedness of parsing is a theorem, not an implementation choice. D3 reflects the logical order of derivation --- axiom precedes terminals. D4 follows from the encoder-decoder architecture of any communication system. D5 is bounded by Gold's theorem. D6 is formalized by the surprisal framework.

The asymmetry does not disappear with technological progress. Large language models displace the analytical cost rather than eliminating it (§1.3). Training a language model IS a massive act of analysis: compressing a corpus into a parametric representation. The \(\mathcal{O}(n)\) generation cost of autoregressive sampling is paid for by the \(\mathcal{O}(N \cdot E)\) training cost, where \(N\) is the corpus size and \(E\) the number of epochs. \emph{Intelligent generation always presupposes prior analysis.}

\subsection{Why bidirectionality has not transferred}

The review of bidirectional systems (§3.3) reveals a striking pattern: bidirectional grammar formalisms have been available since the 1970s (DCG, FST, Q-systems) and refined over five decades (GF, Amalia, grammar inversion). Yet this technology has not transferred to most domain-specific applications.

We propose two hypotheses.

\textbf{Hypothesis 1: The declarativity prerequisite.} Appelt (1987) argued that bidirectionality requires the grammar to be ``represented declaratively.'' Most domain-specific formalisms are procedural --- they embed the processing direction in the grammar itself (e.g., a shape grammar specifies how to \emph{produce} shapes, not how to \emph{recognize} them). Procedural grammars resist inversion because the control structure is inseparable from the linguistic content.

\textbf{Hypothesis 2: The hidden cost.} Bidirectionality has costs not always visible at design time. KAMP was ``computationally impractical.'' Strzalkowski's inverted grammars were less efficient. GF achieves bidirectionality at scale but requires a specific architectural commitment (abstract/concrete syntax separation) that most systems lack. The benefit of bidirectional error detection (+18\%, Goodman \& Bond, 2009) may not justify the engineering effort in domains where only one direction is routinely needed.

These hypotheses are speculative; we offer them as directions for future work. A rigorous test would require identifying systems that \emph{could} have adopted bidirectionality but chose not to, and understanding why.

\subsection{Implications for system design}

\textbf{Benchmarking.} Systems should distinguish between \emph{unconstrained} generation (\(\mathcal{O}(n)\), trivial), \emph{constrained} generation (potentially NP-hard), and recognition. Current benchmarks often conflate these modes.

\textbf{Evaluation metrics.} Surprisal (D6) offers a \emph{unified metric} for the temporal dimension. By measuring average surprisal per token, one can quantify the analytical work a system performs --- whether a parser (explicit analysis) or a language model (implicit analysis encoded in parameters).

\textbf{Architecture choice.} Designers face a choice along D3: design for generation (top-down, intention → surface), recognition (surface → structure), or both? The NLP evidence suggests that bidirectionality, while costly, provides significant advantages for grammar quality.

\subsection{Connections to domain-specific applications}

The six dimensions are \emph{domain-independent} --- they arise from the structure of formal grammars, not from their content. This generality suggests that the asymmetry manifests wherever formal grammars are used.

\textbf{Music} is an ``ideal laboratory'' for the asymmetry: composition (generation) and analysis are established practices with centuries of theoretical tradition. A companion paper (author, in preparation) instantiates all six dimensions on musical systems, adding a seventh dimension specific to the musical domain.

\textbf{Bioinformatics} uses stochastic CFGs (SCFGs) for RNA secondary structure prediction --- a recognition problem. The generation direction (designing RNA sequences with specified structures) is \emph{inverse folding}, known to be NP-hard (\citealt{bonnet2020rna}). The asymmetry manifests as the gap between folding (recognition, polynomial) and design (constrained generation, NP-hard).

\textbf{Programming languages} embody the asymmetry in the gap between compilation (recognition: source → AST → IR) and decompilation (generation: binary → source). Decompilation is possible but produces structurally different code --- a manifestation of D2 (multiple source programs compile to the same binary).

\subsection{The LLM challenge: one model, two directions}

Large language models present an apparent counter-example to the asymmetry thesis. A single autoregressive model (GPT, LLaMA) serves simultaneously as a \textbf{generator} (sampling tokens from \(P(w_i \mid w_1 \cdots w_{i-1})\)) and as a \textbf{recognizer} (computing the probability or perplexity of a given string). The same architecture, the same weights --- only the mode of use changes. Does this unification refute the asymmetry?

We argue it does not, for three reasons.

First, the \textbf{training phase} remains a massive act of analysis: compressing a corpus of \(N\) tokens into \(\theta\) parameters is a recognition/compression operation. The \(\mathcal{O}(n)\) generation cost at inference time is paid for by an \(\mathcal{O}(N \cdot E)\) training cost. The asymmetry is displaced, not eliminated.

Second, the LLM as recognizer is \textbf{fundamentally limited}. It computes \(P(w)\) --- a probabilistic judgment --- but cannot produce a structural description (parse tree, derivation). An LLM can assign high probability to ``the cat sat on the mat'' but cannot identify PP-attachment or assign a phrase-structure tree. Recognition in the formal language sense requires more than a probability score.

Third, \textbf{constrained generation} with LLMs (format compliance, factual accuracy, safety) reintroduces the asymmetry at the output level: generating under constraints is harder than unconstrained sampling, confirming D1.

The LLM case thus illustrates our central thesis: the asymmetry can be \emph{architecturally unified} (same model) while remaining \emph{operationally present} (different costs, different capabilities, different failure modes in each direction). The displacement of analysis from runtime to training time is perhaps the most dramatic contemporary manifestation of the asymmetry.

\section{Conclusion}

The generation-recognition asymmetry in formal grammars is a structural, multidimensional, and irreducible phenomenon. This paper has identified six independent dimensions:

\begin{enumerate}
\def\labelenumi{\arabic{enumi}.}

\item
  \textbf{Computational} (D1): example-generation and recognition both span \(\mathcal{O}(n)\) to \(\notin \mathsf{R}\), but with a \textbf{differential coupling} --- example-generation is mainly sensitive to the task specification (NP-complete under semantic constraint, \citealt{barton1987complexity}), recognition to the grammar class (\(\Theta(n)\) for \(k=3\), PSPACE-complete for \(k=1\), undecidable for \(k=0\)). The asymmetry changes nature with \(k\): quantitative for \(k\in\{2,2^+\}\), structural for \(k=1\), maximal for \(k=0\) (cf.\ §4.1, Table~\ref{tab:d1-membership}).
\item
  \textbf{Ambiguity} (D2): generation is a function (single-valued); parsing is a relation (set-valued), with \(C_n \sim 4^n / n^{3/2}\sqrt{\pi}\) possible parse trees, and some languages are inherently ambiguous (\citealt{parikh1966context}).
\item
  \textbf{Directionality} (D3): generation is invariably top-down (for phrase-structure grammars); parsing has a degree of directional freedom (LL, LR, Earley, CYK) absent from generation.
\item
  \textbf{Information} (D4): the generator knows the source (\(H(X|X) = 0\)); the parser infers from the surface (\(H(X|Y) > 0\)).
\item
  \textbf{Inference} (D5): for fixed grammar class and minimal tasks, gen \textless{} recog \textless{} inference --- bounded by Gold's theorem. Inference can be understood as the extreme of a knowledge continuum whose formal characterization requires further investigation.
\item
  \textbf{Temporality} (D6): the generator creates the future (\(S = 0\)); the parser is expectation-based (\(S > 0\), Levy 2008).
\end{enumerate}

Two of these --- directionality (D3) and temporality (D6) --- have not, to our knowledge, been explicitly identified as structural dimensions within a unified taxonomy of the generation-recognition asymmetry in formal language theory, despite extensive study of their underlying aspects (Russell-Carroll-Warwick 1990; Hale 2001; Levy 2008; and the \emph{P-chain} framework in psycholinguistics, Dell \& Chang 2013).

We have shown that the common characterization ``generation is easy, parsing is hard'' is misleading: the real asymmetry is \emph{structural} --- parsing is always constrained while generation may or may not be.

Bidirectional systems have been available in NLP for over fifty years, yet have not transferred to most domain-specific applications. We proposed two hypotheses: the declarativity prerequisite and the hidden cost of bidirectionality. Large language models architecturally unify generation and recognition but operationally preserve the asymmetry by displacing analysis from runtime to training time.

Although the comprehension/production/acquisition triad has been treated as a unified framework in psycholinguistics (Dell-Chang 2013 P-chain; Chater-Manning 2006), no prior work in formal language theory had treated this asymmetry as a structural dimensional framework. This paper fills that gap at the level of formal language theory, providing a framework for the systematic comparison of formal systems across domains. A companion paper instantiates this framework on musical grammars, where the asymmetry between composition and analysis has both ancient roots and contemporary relevance.

\appendix
\begin{landscape}
\section*{Appendix A --- Summary of the Six Dimensions}
\begin{longtable}[]{@{}
  >{\centering\arraybackslash}p{(\columnwidth - 12\tabcolsep) * \real{0.04}}
  >{\raggedright\arraybackslash}p{(\columnwidth - 12\tabcolsep) * \real{0.12}}
  >{\raggedright\arraybackslash}p{(\columnwidth - 12\tabcolsep) * \real{0.1}}
  >{\raggedright\arraybackslash}p{(\columnwidth - 12\tabcolsep) * \real{0.1}}
  >{\raggedright\arraybackslash}p{(\columnwidth - 12\tabcolsep) * \real{0.15}}
  >{\raggedright\arraybackslash}p{(\columnwidth - 12\tabcolsep) * \real{0.2}}
  >{\raggedright\arraybackslash}p{(\columnwidth - 12\tabcolsep) * \real{0.25}}@{}}
\toprule\noalign{}
\begin{minipage}[b]{\linewidth}\centering
Dim
\end{minipage} & \begin{minipage}[b]{\linewidth}\raggedright
Name
\end{minipage} & \begin{minipage}[b]{\linewidth}\raggedright
Gen
\end{minipage} & \begin{minipage}[b]{\linewidth}\raggedright
Parse
\end{minipage} & \begin{minipage}[b]{\linewidth}\raggedright
Key ref
\end{minipage} & \begin{minipage}[b]{\linewidth}\raggedright
Counter-argument
\end{minipage} & \begin{minipage}[b]{\linewidth}\raggedright
Resolution
\end{minipage} \\
\midrule\noalign{}
\endhead
\bottomrule\noalign{}
\endlastfoot
D1 & Computational & \(\mathcal{O}(n)\) to \(\notin \mathsf{R}\) & \(\mathcal{O}(n)\) to \(\notin \mathsf{R}\) & Younger 1967, Valiant 1975, Satta 1994, Savitch 1970, Barton et al.\ 1987 & Gen NP-hard under semantic constraint & Differential coupling: gen ← task spec., recog ← grammar class; asymmetry changes nature with \(k\) (quantitative \(k\in\{2,2^+\}\), structural \(k=1\), maximal \(k=0\)) \\
D2 & Ambiguity & Function (single-valued) & Relation (multivalued), \(C_n \sim 4^n\); inherently ambiguous langs. & Parikh 1966, Brabrand 2007 & Deterministic langs. unambig. & DCFLs exclude natural langs.; inherent ambiguity is a theorem \\
D3 & Directionality & Top-down (phrase-struct.) & Free choice (LL, LR, \ldots) & Shieber 1988, 1990 & L-systems, shape grammars & Scoped to phrase-struct.; others confirm asym. \\
D4 & Information & \(H(X \mid X) = 0\) & \(H(X \mid Y) > 0\) & Shannon 1948 & Bits-Back exploitable & Asymmetry = resource \\
D5 & Inference & gen \textless{} recog \textless{} infer (canonical \(k\!=\!2\)) & Knowledge continuum & Gold 1967 & PCFG escapes Gold & Still ``hard as RSA'' (\citealt{kearns1994cryptographic}) \\
D6 & Temporality & Causal: \(S = 0\) & Expectation-based: \(S > 0\) & Hale 2001, Levy 2008 & Stochastic gen \(S > 0\) & Chosen vs.~imposed uncertainty \\
\end{longtable}
\end{landscape}

% Bibliography — .bbl inlined for arXiv compilation
\bibliographystyle{elsarticle-num-names}
\bibliography{manuscript}

\end{document}